\documentclass[letterpaper, 10 pt, conference]{ieeeconf}  

\IEEEoverridecommandlockouts                              

\usepackage{amsmath} 
\usepackage{amssymb}  
\usepackage{booktabs}
\usepackage{multirow}
\usepackage{array}
\usepackage{pifont}
\usepackage[table]{xcolor}
\usepackage{threeparttable}
\usepackage{adjustbox}
\usepackage{pgfplots}
\pgfplotsset{compat=1.17}
\usepgfplotslibrary{statistics}  
\usepackage{listings}
\usepackage{url}
\usepackage{stfloats}
\usepackage{caption}
\usepackage{graphicx}

\title{\LARGE \bf
HumanoidVLN: A Physics-Grounded Simulator and Benchmark for Vision-Language Navigation Across Diverse Humanoid Embodiments
}
\author{Quan-Dung Pham$^{1, *}$, Anh Dao$^{1, *}$, The-Anh Nguyen$^{1, *}$, Minh Nguyen-Dinh$^{1, *}$, Phuong~Nam Dang$^{1, *}$, \\ Tri Pham$^{1}$, Hung Tran$^{1}$, Bach Dao$^{1}$, Tuyen P. Le$^{1}$, Truong Nguyen$^{1}$, Quan Nguyen$^{2}$
\thanks{$^{1}$VinMotion, Inc., Vietnam}
\thanks{$^{2}$University of Southern California, USA}
\thanks{$^{*}$Equal contribution}
}

\begin{document}

\IEEEaftertitletext{%
    \vspace{4pt}%
    \begin{center}
        \includegraphics[width=0.85\textwidth]{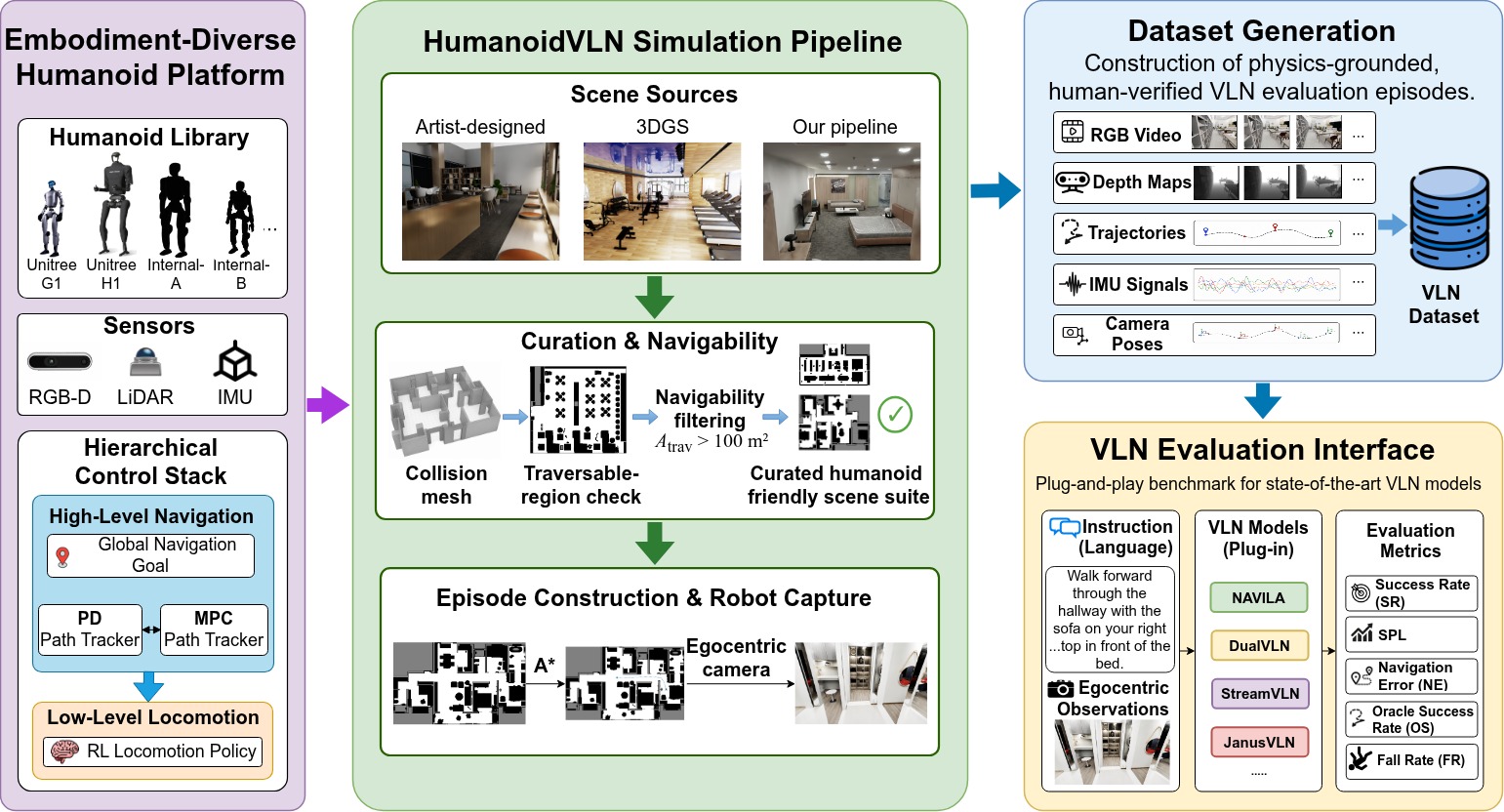}
        \captionof{figure}{Overview of the \textbf{HumanoidVLN} physics-grounded benchmark pipeline, from embodiment-diverse humanoid platforms and hierarchical control to scene curation, episode construction, multimodal dataset generation, and plug-and-play VLN evaluation.}
        \label{fig:simulation_overview}
    \end{center}
    \vspace{4pt}%
}
\maketitle
\thispagestyle{empty}
\pagestyle{empty}


\begin{abstract}
Vision-Language Navigation (VLN) for humanoid robots presents unique
challenges that existing benchmarks fail to address: bipedal locomotion imposes physical constraints absent from wheeled or idealized agents, humanoid morphologies vary significantly across platforms, and egocentric observations are distorted by locomotion-induced camera dynamics. We present \textbf{HumanoidVLN}, a physics-grounded simulator and benchmark for VLN across diverse humanoid embodiments. Built on NVIDIA Isaac Sim, our platform supports an extensible set of humanoid configurations — demonstrated on four robots (Unitree G1, Unitree H1, Internal-A, Internal-B) spanning 10–12 lower-body DoF and heights from 1.17\,m to 1.80\,m — via a hierarchical control stack combining a Reinforcement Learning locomotion policy with interchangeable PD or MPC path trackers. New robots and VLN models can be integrated with minimal effort, and we demonstrate compatibility with NaVILA, DualVLN, StreamVLN, and JanusVLN. Environments are drawn from artist-designed scenes and 3D Gaussian Splatting reconstructions,
filtered for navigability areas exceeding 100\,m\textsuperscript{2}. Navigation instructions are generated by a \textit{Dual Generator-Reviewer + Paraphraser} Multi-Agent Annotation with human-in-the-loop verification, yielding 933 collision-aware reference episodes, each paired with one spatially grounded fine-grained instruction and three coarse-grained stylistic variants (Formal, Natural, Casual). Across four models and four humanoid embodiments, JanusVLN achieves the highest mean SR of 43.55\% and nDTW of 48.38. In a 20-episode sim--real pilot with DualVLN and Unitree G1, navigation errors are strongly correlated ($r=0.935$), with a mean absolute difference of 0.68m and mean trajectory similarity of $0.782\pm0.188$ nDTW. These results highlight the interaction between VLN models, controllers, and humanoid embodiments under physical execution. \textit{Code, benchmark, and data will be released upon acceptance at} \url{https://humanoid-vln.github.io/}.
\end{abstract}

\section{INTRODUCTION}
 
The past decade has witnessed remarkable advancements in humanoid
robotics~\cite{feng2014optimization,gu2025humanoid}, driven by progress in perception, control, and machine learning. In parallel,
vision-language navigation (VLN)~\cite{anderson2018vision,narayan2019collaborative}
has emerged as a key capability for embodied agents to interpret and
execute natural language instructions in complex environments. The sim-to-real gap is particularly pronounced for humanoid platforms, where discrepancies in dynamics, sensing, and embodiment constraints can substantially affect navigation performance. We identify three critical gaps and introduce targeted solutions.
 
\vspace{2pt}
\noindent\textbf{Gap 1: Humanoid-Aware Simulation.}
Existing VLN simulators model robot motion through kinematic teleportation, bypassing the physical constraints of bipedal locomotion. Recent Isaac Sim-based works address physical realism but treat the humanoid as one of several interchangeable robot types, sharing a single control proxy across morphologies. Humanoid robots are uniquely diverse --- lower-body DOF ranges from 10 to 12, heights from 1.17\,m to 1.8\,m --- and these differences directly impact gait stability, step reachability, and CoM dynamics. We develop an Isaac Sim platform with a hierarchical control architecture: an RL policy governs low-level locomotion while interchangeable Proportional-Derivative (PD) or Model Predictive Control (MPC) controllers serve as high-level path trackers, grounding navigation in actual bipedal dynamics across four humanoid morphologies.
 
\vspace{2pt}
\noindent\textbf{Gap 2: Navigability-Curated Scene Environments.}
Standard VLN benchmarks rely on Matterport3D~\cite{chang2017matterport3d},
offering limited realism. More recent works, such as GRScenes~\cite{grutopia} and SAGE-3D~\cite{miao2025towards} improve scene quality but scale scene \emph{count} without curating for
\emph{navigability}. Humanoid robots require large, obstacle-sparse
traversable areas; scenes lacking this criterion introduce topological bottlenecks infeasible for bipedal navigation. We curate scenes with manually verified traversable areas of at least 100\,m\textsuperscript{2}, a practical threshold used to exclude small or highly constrained environments, and employ a Real2Sim pipeline based on 3D Gaussian Splatting (3DGS)~\cite{kerbl20233d} to construct simulation-ready environments without requiring fully artist-authored assets~\cite{miao2025towards}.

 
\vspace{2pt}
\noindent\textbf{Gap 3: Realistic, Grounded Instruction Data.}
Existing benchmarks render videos under idealized conditions --- stable camera, uniform lighting --- absent in real humanoid operation, where bipedal gait induces camera shake and dynamic lighting variation. Fully automated VLM-based annotation~\cite{lin2025vlnverse} scales well but
suffers from spatial hallucinations; manual annotation is reliable but prohibitively costly. We collect egocentric video from our humanoid simulator and propose a \textit{Dual Generator-Reviewer + Paraphraser} Multi-Agent Annotation (MAA) that iteratively generates, verifies, and paraphrases instructions, augmented by a human-in-the-loop verification pass.

\vspace{4pt}
\noindent The primary contributions of this work are:
\begin{itemize}
    \item \textbf{Embodiment-Diverse Humanoid Simulation Platform:} An Isaac Sim framework supporting Unitree G1 (12 DoF, 1.32\,m), Unitree H1 (10 DoF, 1.80\,m), Internal-A (12 DoF, 1.61\,m), and Internal-B (12 DoF, 1.17\,m) via a hierarchical RL-plus-path-tracker control stack, with a plug-and-play interface for additional robots and VLN
    models. 
 
    \item \textbf{Navigability-Curated Real2Sim Scene Suite:} 87 high-fidelity 3D environments ($\geq$100\,m\textsuperscript{2} navigable area) spanning 17 indoor classes across 6 domains, drawn from artist-designed and 3DGS reconstructed sources.
 
    \item \textbf{MAA-Augmented Instruction Dataset:} 933 episodes each with one spatially grounded fine-grained instruction and three coarse-grained stylistic variants, generated by a \textit{Dual Generator-Reviewer + Paraphraser} MAA and verified by human annotators.
 
    \item \textbf{Cross-Embodiment Physics Evaluation:}
    A matched zero-shot evaluation of four representative VLN models
    across four humanoid configurations under physics-based execution,
    measuring navigation accuracy, path fidelity, and Fall Rate.
\end{itemize}

\section{RELATED WORK}
 
\subsection{Simulation Platforms for Embodied Navigation}
Habitat-Sim~\cite{savva2019habitat} dominates embodied VLN but models
motion through kinematic stepping that bypasses physical locomotion dynamics. AI2-THOR~\cite{kolve2017ai2} and AirSim~\cite{shah2017airsim} target manipulation and aerial navigation respectively, sharing the same limitation. NVIDIA Isaac Sim provides full rigid-body dynamics and high-DOF articulation. VLN-PE~\cite{wang2025rethinking} first leveraged Isaac Sim for VLN, exposing locomotion-induced failure modes across multiple robot types. VLNVerse~\cite{lin2025vlnverse} scaled this to large scene collections with full-kinematics simulation. However, both treat the humanoid as one of several interchangeable morphologies without specializing control, scene selection, or metrics to bipedal constraints. HumanoidVLN is designed exclusively for humanoids, with each of four morphologies driven by its own RL locomotion policy beneath interchangeable PD/MPC path trackers, ensuring every evaluated trajectory is physically executable by the specific robot under test.
 
\noindent\textbf{3D Scene Reconstruction.}
3DGS~\cite{kerbl20233d} enables fast and efficient 3D reconstruction, rendering high-quality novel views at simulation-compatible frame rates. Isaac Sim 5.1 integrates native rendering of 3D Gaussians ~\cite{loccoz20243dgrt,wu20253dgut}, layered on top of collision meshes. SAGE-3D~\cite{miao2025towards} applies 3DGS to VLN scenes but omits navigability filtering and requires artist-authored meshes --- both addressed in our pipeline.

\subsection{Vision-Language Navigation Benchmarks}
Early VLN benchmarks established core task formulations on Matterport3D~\cite{chang2017matterport3d}: R2R~\cite{anderson2018vision} defined canonical path-following on panoramic viewpoint graphs, R4R~\cite{jain2019stay} extended it with compositional trajectories, and RxR~\cite{ku2020room} introduced multilingual, spatiotemporally dense annotations. Subsequent works broadened task scope: REVERIE~\cite{qi2020reverie} and SOON~\cite{zhu2021soon} added object-grounded navigation, while CVDN~\cite{thomason2020vision} and VNLA~\cite{nguyen2019vision} introduced interactive and dialog-driven settings. Toward physical continuity, VLN-CE~\cite{krantz2020beyond} moved to continuous action spaces, LH-VLN~\cite{song2025towards} targeted long-horizon multi-stage tasks,  HA-VLN~\cite{li2024human} integrated dynamic human activities. GSA-R2R~\cite{hong2025general} improved visual fidelity via 3DGS reconstruction. Despite these advances, none account for the morphological constraints of bipedal humanoids, where gait stability and CoM dynamics fundamentally alter trajectory feasibility.
 
 
\providecommand{\cmark}{\textcolor{teal!70!black}{\ding{51}}}
\providecommand{\cmark}{\textcolor{teal!70!black}{\ding{51}}}
\providecommand{\na}{\textcolor{gray}{--}}
\providecommand{\ours}{\textbf{Ours}}
 
\begin{table}[h]
    \centering
    \caption{Comparison of representative VLN benchmarks. \textbf{Hum.}: a simulator purpose-built for bipedal humanoid robots. \textbf{Scene}: A=Artist-designed, GS=3D Gaussian Splatting. \textbf{Instr.}: H=Human, VLM=automated VLM, MAA+H=Multi-Agent Annotation with human-in-the-loop verification.}
    \label{tab:vln_benchmark_comparison}
    \setlength{\tabcolsep}{3pt}
    \renewcommand{\arraystretch}{1.1}
    \begin{threeparttable}
    \begin{adjustbox}{width=\columnwidth}
    \scriptsize
    \begin{tabular}{l l c c c c c}
        \toprule
        \textbf{Dataset} & \textbf{Sim.} & \textbf{Hum.} & \textbf{DoF} & \textbf{Scene} & \textbf{Instr.} & \textbf{Action} \\
        \midrule
        R2R (2018)       & MP3D    & \na & \na    & A      & H      & Graph    \\
        RxR (2020)       & MP3D    & \na & \na    & A      & H      & Graph    \\
        REVERIE (2020)   & MP3D    & \na & \na    & A      & H      & Graph    \\
        VLN-CE (2020)    & Habitat & \na & \na    & A      & H      & Discrete \\
        ALFRED (2020)    & THOR    & \na & \na    & A      & H      & Discrete \\
        AerialVLN (2023) & AirSim  & \na & \na    & A      & H      & Discrete \\
        LH-VLN (2025)    & Habitat & \na & \na    & A      & VLM    & Graph    \\
        GSA-R2R (2025)   & Habitat & \na & \na    & GS     & VLM    & Graph    \\
        VLN-PE (2025)    & Isaac   & \na & \na    & A      & VLM    & Hybrid   \\
        \rowcolor{blue!8}
        VLNVerse (2025)  & Isaac   & \na & \na    & A      & VLM    & Hybrid   \\
        \midrule
        \rowcolor{cyan!12}
        \ours{} (2026)   & Isaac   & \cmark & 10--12 & A+GS & MAA+H  & Hybrid   \\
        \bottomrule
    \end{tabular}
    \end{adjustbox}
    \begin{tablenotes}
        \scriptsize
        \item \textbf{DoF}: Unitree G1 12\,DoF/1.32\,m; H1 10\,DoF/1.80\,m; Int-A 12\,DoF/1.61\,m; Int-B 12\,DoF/1.17\,m. Full details omitted for double-blind review.
    \end{tablenotes}
    \end{threeparttable}
\end{table}
Table~\ref{tab:vln_benchmark_comparison} summarizes key dimensions across representative benchmarks. No existing work jointly addresses humanoid embodiment diversity, navigability-curated scenes, and reliable grounded instruction generation --- the three axes that define our contribution.

\section{THE HUMANOIDVLN PLATFORM}
 
\subsection{Supported Embodiments}
\begin{table}[t]
    \centering
    \caption{Humanoid embodiments supported by \textbf{HumanoidVLN}.
    Internal platform details omitted for double-blind review.}
    \label{tab:robots}
    \setlength{\tabcolsep}{2.pt}
    \renewcommand{\arraystretch}{1.2}
    \small
    \begin{tabular}{l c c c}
        \toprule
        \textbf{Robot} & \textbf{Lower-body DoF} & \textbf{Height (m)} & \textbf{Camera height (m)} \\
        \midrule
        Unitree G1   & 12 & 1.32 & 1.25 \\
        Unitree H1   & 10 & 1.80 & 1.72 \\
        Internal-A   & 12 & 1.61 & 1.54 \\
        Internal-B   & 12 & 1.17 & 1.11 \\
        \bottomrule
    \end{tabular}
\end{table}
 
The platform is organized around four humanoid configurations (Table~\ref{tab:robots}) chosen to span the embodiment diversity of current bipedal robots in height, lower-body DoF, and camera height. 
Each configuration loads independently, enabling controlled cross-embodiment evaluation under identical environmental conditions.
 
 
\subsection{Hierarchical Control Architecture}
Each robot is governed by a two-level control hierarchy, as illustrated in Fig.~\ref{fig:simulation_overview}. 
At the low level, a per-embodiment Reinforcement Learning (RL) locomotion policy commands joint torques to produce stable bipedal gaits while respecting per-morphology joint limits and center-of-mass (CoM) dynamics. At the high level, either a PD controller or an MPC path tracker converts a high-level navigation plan into velocity and heading references for the RL policy. Tracker selection is matched to each model's action space: the MPC tracker is used for continuous-action models, and the PD tracker for discrete-action models. Grounding path following in this hierarchy rather than kinematic teleportation ensures that every evaluated trajectory is subject to the true physical constraints of the specific morphology under test.
 
\subsection{VLN Model Interface and Data Recording}
The platform serves a dual purpose. 
First, it records egocentric observation streams --- RGB video, depth, and IMU --- directly from the walking robot, capturing the camera shake and dynamic lighting induced by bipedal gait; these streams form the perceptual input of our instruction pipeline (Sec.~\ref{sec:instructions}). Second, it exposes a single observation-action interface: a VLN model receives the head-camera RGB stream and the instruction, and emits actions that the platform translates into tracker references. Integrating a new model requires implementing only this interface, with no model-side modification. We demonstrate compatibility with NaVILA, DualVLN, StreamVLN, and JanusVLN. Integrating a new robot requires an URDF/USD description and a trained locomotion policy.

\section{REAL2SIM SCENE PIPELINE}
 
\subsection{Scene Sources and Navigability Curation}
HumanoidVLN environments come from two complementary sources:
artist-designed indoor scenes and 3DGS reconstructions.
Because humanoids require large, obstacle-sparse traversable regions, we retain only scenes whose traversable floor area --- verified against the extracted collision mesh --- exceeds 100\,m\textsuperscript{2}. The final suite contains 87 scenes spanning 17 indoor classes across 6 application domains (Residential, Food \& retail, Culture \& leisure,
Workplace \& education, Healthcare, Fitness), with a median navigable area of 266\,m\textsuperscript{2} and a mean of 387\,m\textsuperscript{2}.
 
\subsection{3DGS Reconstruction}
While SAGE-3D~\cite{miao2025towards} relies on synthetic 3D assets, we develop a customized 3DGS reconstruction pipeline that allows us to produce high-quality splats and collision meshes without the need for artist-authored scenes. Moreover, our pipeline directly consumes multi-view captures of the target environment as training data, producing photorealistic scenes as opposed to images rendered from synthetic assets.

We train splats using the open-source \texttt{gsplat}~\cite{ye2025gsplat} library, with 3D Gaussians as the main radiance field primitives. While recent works~\cite{Huang2DGS2024, svraster2025} propose to use alternative scene primitives for better mesh extraction, 3D Gaussians remain the first-class scene assets supported by Isaac Sim and other emerging simulation platforms~\cite{xia2026habitatgs}. To extract high-fidelity meshes post-training for collision reasoning, we modify the training pipeline of \texttt{gsplat} to render unbiased depth~\cite{Chen2025pgsr} and enforce depth-normal consistency~\cite{Huang2DGS2024}. Collision meshes are extracted post-training via TSDF fusion~\cite{Newcombe2011tsdf} of rendered depth maps and packed into the same USDZ scene with the trained Gaussians.

We register multi-view captures and bootstrap Gaussian training from COLMAP ~\cite{schoenberger2016sfm,schoenberger2016mvs}. To align COLMAP's OpenCV camera pose convention ($-Y$-up, $+Z$-forward) with Isaac Sim's frame ($+Z$-up, $+Y$-forward), we apply a fixed axis-switching transform when parsing COLMAP's sparse database.

 
\subsection{Physics-Aware Episode Sampling}
For each scene, we slice the 3D collision mesh at the robot's body height to form a 2D occupancy map capturing overhead obstacles, and dilate it by the circumscribed footprint radius of the largest supported robot. Start--goal pairs are sampled randomly and connected by A$^*$ paths on this map. The robot executes each path under its full control stack while the platform records egocentric RGB/depth/IMU streams; paths that cannot be completed stably are resampled. By construction, every episode's video contains genuine locomotion-induced camera dynamics.

\section{INSTRUCTION GENERATION}
\label{sec:instructions}
 
\subsection{Multi-Agent Annotation with Evidence-Scoped Roles}
\label{subsec:maa}

Single-pass VLM annotation can produce incorrect turns, unsupported landmarks, and temporally misaligned actions. 
We therefore separate instruction generation from evidence-grounded verification, following the general Describer--Verifier--Synthesizer paradigm of VLNVerse~\cite{lin2025vlnverse}. 
Our design differs in three respects: the generators do not access ground-truth geometric priors, verification combines deterministic geometric checks with VLM reasoning, and the resulting instructions undergo final human correction.

 
\begin{figure*}[t]
    \centering
    \includegraphics[width=0.92\textwidth]{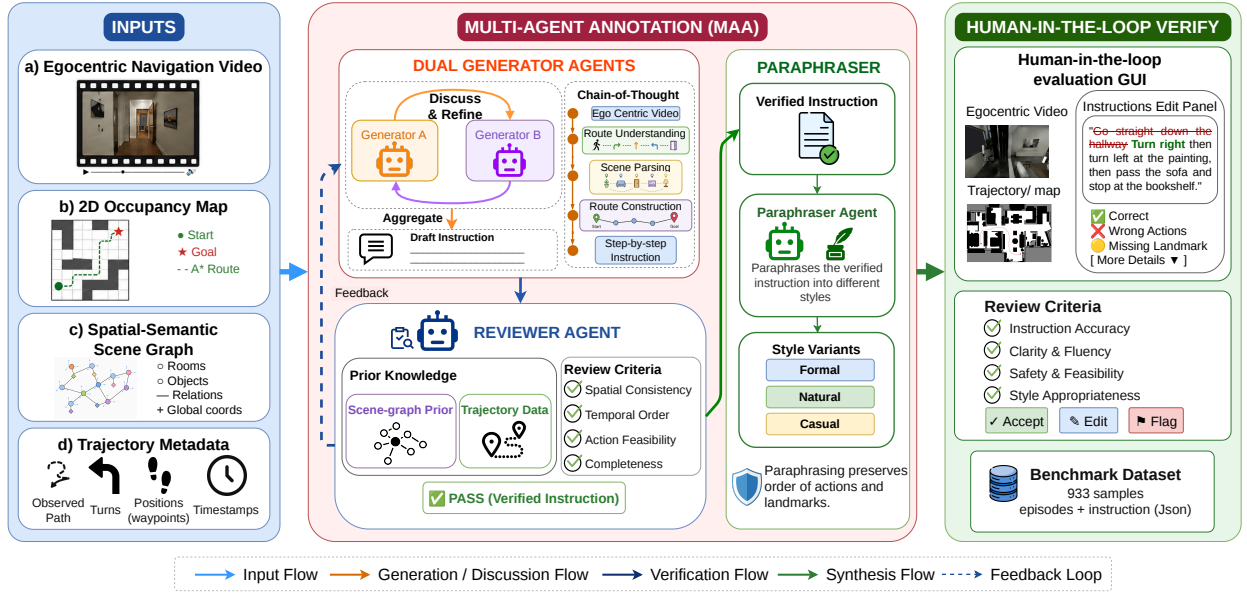}
    \caption{The proposed \textit{Multi-Agent Annotation (MAA)} framework.
    Two generator agents independently derive a route graph from the egocentric keyframe sequence alone; their graphs are reconciled, and residual contradictions are forwarded to a reviewer agent that verifies against scene-graph, trajectory, and occupancy priors.
    The verified instruction is paraphrased into three stylistic variants and corrected by human annotators to produce the benchmark dataset.}
    \label{fig:pipeline}
\end{figure*}

\noindent\textbf{Goal grounding and route generation.}
Qwen3-VL-30B-A3B \cite{bai2025qwen3} identifies a candidate goal landmark and stopping condition from the terminal frames and nearby scene-graph objects. 
The same goal condition is provided to both generators, while all remaining route content must be inferred from the egocentric keyframes alone.
Gemma-4-31B-it \cite{gemmateam2026gemma4} and InternVL3.5-38B~\cite{wang2025internvl3} independently perform a structured route decomposition consisting of motion interpretation, landmark grounding, route construction, and instruction realization.

Each generator produces an intermediate route graph
\[
R=\langle(a_i,\ell_i,s_i,o_i,m_i)\rangle_{i=1}^{n},
\]
where $a_i$, $\ell_i$, $s_i$, $o_i$, and $m_i$ denote the ordered action, landmark, side-of-path relation, ordinal, and turn magnitude, respectively. 
Comparing structured routes rather than surface text allows navigationally equivalent instructions to be matched despite differences in wording. 
The heterogeneous generators are intended to provide complementary route interpretations and reduce correlated model-specific errors, but do not replace subsequent verification.

\noindent\textbf{Reconciliation and verification.}
The two route graphs are compared using navigation-critical attributes, including turn direction, normalized landmark identity, and side-of-path relations, ordinals, and step order. 
Non-conflicting additions are merged, while localized contradictions are forwarded to the reviewer. 
The reviewer, Qwen3-VL-30B-A3B \cite{bai2025qwen3}, verifies the reconciled route using trajectory metadata, the A$^{*}$ route over the occupancy map, and a spatial--semantic scene graph. 
Per-frame semantic masks restrict the scene-graph evidence to objects visible along the executed trajectory, preventing verification against landmarks that the generators could not observe. 
Deterministic checks validate geometrically decidable properties such as turn direction, temporal order, landmark identity, and spatial relations, while the VLM reviewer handles completeness, referring-expression quality, and residual ambiguity.

\noindent\textbf{Instruction realization and paraphrasing.}
The verified route is converted into one fine-grained instruction.
GPT-5.5 then receives only the verified text and produces Formal, Natural, and Casual variants following~\cite{lin2025vlnverse} ($T=1.0$).
Each variant is re-parsed into the route representation and accepted only when its navigation-critical attributes match those of the verified route. 
Using different model families for generation, review, and paraphrasing is intended to reduce correlated errors and self-preference; final human verification remains necessary.

\subsection{Human-in-the-Loop Verification}
\label{subsec:human_verification}
All generated instructions are reviewed by a pool of three trained annotators. 
For each episode, one annotator examines the complete egocentric video and route visualization and verifies the action order, landmarks, spatial relations, and stopping conditions. 
Residual errors are corrected, while ambiguous or infeasible instructions are flagged
A randomly selected $20\%$ of the episodes is independently reviewed by a second annotator, with disagreements resolved by the third.
 
\subsection{Benchmark Composition}
HumanoidVLN comprises \textbf{933 episodes} across 87 scenes. 
Each episode is stored in a standardized JSON schema that includes landmarks, a route summary, one spatially grounded, fine-grained instruction, and three stylistic variants (Formal, Natural, and Casual). 
The benchmark is used solely for zero-shot evaluation: all scenes and episodes form a single evaluation set, and no evaluated model is trained or fine-tuned on HumanoidVLN. 
For evaluation, one instruction per episode is selected using a fixed, approximately balanced assignment across the four styles and shared across all model--robot configurations.


\section{EXPERIMENTS}

\subsection{Setup}
 
\noindent\textbf{Models and Protocol.} We evaluate four models representing distinct architectural directions: NaVILA~\cite{cheng2024navila}, built on the NVILA foundation model for discrete-action navigation; StreamVLN~\cite{wei2025streamvln}, which encodes continuous egocentric
video streams particularly suited to humanoid camera dynamics; DualVLN~\cite{wei2025ground}, a dual-system model decoupling semantic reasoning from action prediction with continuous velocity outputs; and JanusVLN~\cite{zeng2025janusvln}, which incorporates explicit 3D spatial
understanding into action decisions. NaVILA, StreamVLN, and JanusVLN operate in a discrete action space (\textit{forward}, \textit{turn left}, \textit{turn right}, \textit{stop}) and are paired with the PD tracker; DualVLN outputs continuous velocity commands and is paired with the MPC tracker. All public checkpoints are evaluated zero-shot on the same 933 episodes without training or fine-tuning on HumanoidVLN. For each episode, one of the four available instructions is selected using a fixed, approximately balanced assignment across styles, shared by all model--robot configurations, and released with the benchmark for reproducibility.

\noindent\textbf{Metrics.} Following standard VLN protocol we report Success Rate (SR; stop within 3.0\,m of goal), Oracle Success Rate (OS), Navigation Error (NE, m), Success weighted by Path Length (SPL), and Normalized Dynamic Time Warping (nDTW; path fidelity independent of task success). 
We additionally report episode-level \textbf{Fall Rate (FR)}, defined as
\begin{equation}
\mathrm{FR}
=
\frac{100}{N}
\sum_{i=1}^{N}
\mathbb{1}[F_i=1],
\end{equation}
where $F_i=1$ if the initially upright robot satisfies at least one fall-detection criterion. Let $\Delta h$ denote the decrease from the nominal upright base height and $H_e$ the nominal height of embodiment $e$: \textbf{T1} (dynamic fall), $\Delta h \geq 0.5H_e$ with downward-speed magnitude above $1.2\,\mathrm{m/s}$; \textbf{T2} (sustained collapse), $\Delta h \geq 0.5H_e$ for at least $2\,\mathrm{s}$; or \textbf{T3} (shallow dynamic fall), $0.35H_e \leq \Delta h < 0.5H_e$ with downward-speed magnitude above $1.5\,\mathrm{m/s}$. The initial-upright requirement and height thresholds exclude normal crouching. Once a fall is detected, the episode is terminated without automatic recovery.

 
\noindent We organize experiments around four questions. \textbf{Q1:} How do state-of-the-art VLN models perform under physically grounded humanoid execution (Sec.~\ref{subsec:main_results})? \textbf{Q2:} How does embodiment configuration affect performance (Sec.~\ref{subsec:main_results})? \textbf{Q3:} Does our 3DGS reconstruction pipeline produce simulation-ready environments of sufficient quality
(Sec.~\ref{subsec:rec_quality})? \textbf{Q4:} Does performance in reconstructed scenes predict real-world performance (Sec.~\ref{subsec:real2sim})?
 
\subsection{Q1 \& Q2: Cross-Embodiment Evaluation}
\label{subsec:main_results}

\definecolor{bestcell}{RGB}{186, 230, 201}    
\definecolor{secondcell}{RGB}{255, 224, 178}  
\newcommand{\best}[1]{\cellcolor{bestcell}\textbf{#1}}
\newcommand{\second}[1]{\cellcolor{secondcell}#1}

\begin{table}[t]
    \centering
    \caption{Zero-shot VLN model performance on \textbf{HumanoidVLN}
    evaluation set across four humanoid embodiments.
    FR is the percentage of episodes containing at least one detected fall, measuring locomotion stability under physical execution.
    Internal platform details omitted for double-blind review.
    Best results are highlighted in green (bold); second-best in amber.}
    \label{tab:main_results}
    \setlength{\tabcolsep}{3pt}
    \renewcommand{\arraystretch}{1.05}
    \begin{adjustbox}{width=\columnwidth}
    \scriptsize
    \begin{tabular}{l l c c c c c c}
        \toprule
        \textbf{Model} & \textbf{Robot} & \textbf{SR}$\uparrow$ &
        \textbf{OS}$\uparrow$ & \textbf{NE}$\downarrow$ &
        \textbf{SPL}$\uparrow$ & \textbf{nDTW}$\uparrow$ & \textbf{FR}$\downarrow$ \\
        \midrule
        \multirow{4}{*}{NaVILA}
            & Unitree G1 & 28.19 & 41.80 & 5.86 & 22.24 & 43.65 & 7.93 \\
            & Unitree H1 & 21.97 & 37.41 & 6.14 & 17.12 & 41.97 & 70.95 \\
            & Internal-A & 31.73 & 56.27 & 5.54 & 15.81 & 31.99 & \best{2.68} \\
            & Internal-B & 26.80 & 42.77 & 5.94 & 21.00 & 37.93 & 6.97 \\
        \midrule
        \multirow{4}{*}{StreamVLN}
            & Unitree G1 & 26.80 & 34.30 & 6.35 & 19.50 & 41.38 & 9.54 \\
            & Unitree H1 & 14.36 & 20.58 & 7.32 & 11.17 & 35.45 & 64.52 \\
            & Internal-A & 31.83 & 40.41 & 6.07 & 19.51 & 39.72 & \best{2.68} \\
            & Internal-B & 21.54 & 29.80 & 7.13 & 11.63 & 30.15 & 7.93 \\
        \midrule
        \multirow{4}{*}{DualVLN}
            & Unitree G1 & 35.05 & 42.55 & 5.73 & 25.83 & 46.67 & 5.47 \\
            & Unitree H1 & 17.68 & 22.19 & 6.80 & 14.70 & 39.17 & 27.12 \\
            & Internal-A & 32.69 & 41.37 & 5.70 & 21.86 & 43.27 & 3.64 \\
            & Internal-B & 30.23 & 40.51 & 5.93 & 23.29 & 44.44 & 7.07 \\
        \midrule
        \multirow{4}{*}{JanusVLN}
            & Unitree G1 & 44.59 & 55.31 & 4.87 & 33.18 & \second{50.08} & 7.40 \\
            & Unitree H1 & 29.37 & 38.48 & 5.80 & 23.24 & 44.93 & 30.76 \\
            & Internal-A & \best{50.38} & \second{63.34} & \best{4.26} & \second{33.56} & \best{50.51} & \second{2.79} \\
            & Internal-B & \second{49.84} & \best{64.31} & \second{4.54} & \best{36.18} & 48.01 & 9.97 \\
        \bottomrule
    \end{tabular}
    \end{adjustbox}
\end{table}
 
Table~\ref{tab:main_results} summarizes results for Q1 and Q2.
 
\noindent\textbf{Q1 --- Overall model performance.}
JanusVLN achieves the highest average SR (43.55\%) and nDTW (48.38), a result consistent with potential benefits from its explicit 3D spatial representation under physical humanoid execution. DualVLN leads the remaining models in nDTW (43.39), indicating stronger path adherence. NaVILA exhibits the largest average OS--SR gap (17.39 points), consistent with less reliable final stopping decisions. StreamVLN records the lowest SR (23.63\%) and nDTW (36.67) among the four evaluated models.
 
\noindent\textbf{Q2 --- Embodiment effects.}
Unitree H1 (10 DoF, 1.80,m) shows the lowest performance across all models, with an average SR of 20.84\% compared with 32--37\% for the other robots. The largest cross-robot SR differences occur for JanusVLN (18.90 points) and DualVLN (14.98 points). Internal-A achieves the highest average SR (36.66\%), whereas G1 achieves the highest average SPL (25.19), showing that robot rankings vary across metrics.

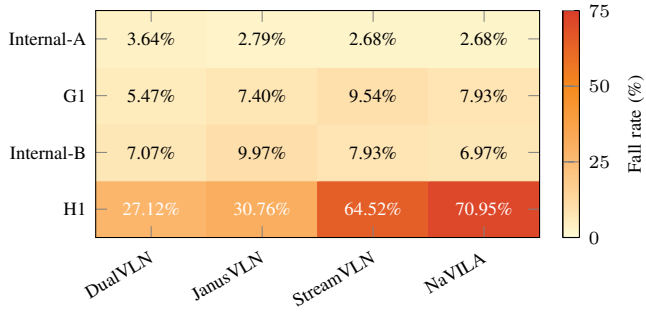
\begin{figure}[t]
\centering
\begin{tikzpicture}
\begin{axis}[
    width=0.68\linewidth, height=3cm,
    colormap={fallrate}{
        color(0)=(yellow!20);
        color(33)=(orange!60);
        color(66)=(orange!90!red);
        color(100)=(red!90!black)
    },
    colorbar,
    colorbar style={
        ylabel={Fall rate (\%)},
        ylabel style={font=\scriptsize},
        tick label style={font=\scriptsize},
        width=0.22cm,
        ymin=0, ymax=75,
        ytick={0,25,50,75},
    },
    point meta min=0,
    point meta max=75,
    xtick={0,1,2,3},
    xticklabels={DualVLN, JanusVLN, StreamVLN, NaVILA},
    x tick label style={
        font=\scriptsize,
        rotate=30,
        anchor=north east,
    },
    ytick={0,1,2,3},
    yticklabels={H1, Internal-B, G1, Internal-A},
    y tick label style={font=\scriptsize},
    xmin=-0.5, xmax=3.5,
    ymin=-0.5, ymax=3.5,
    axis on top,
    scale only axis,
]

\addplot[
    matrix plot*,
    mesh/cols=4,
    point meta=explicit,
] coordinates {
    (0,0) [27.1]  (1,0) [30.8]  (2,0) [64.5]  (3,0) [71.0]
    (0,1) [7.1]   (1,1) [10.0]  (2,1) [7.9]   (3,1) [7.0]
    (0,2) [5.5]   (1,2) [7.4]   (2,2) [9.5]   (3,2) [7.9]
    (0,3) [3.6]   (1,3) [2.8]   (2,3) [2.7]   (3,3) [2.7]
};

\node[font=\scriptsize, text=black] at (axis cs:0,3) {3.64\%};
\node[font=\scriptsize, text=black] at (axis cs:1,3) {2.79\%};
\node[font=\scriptsize, text=black] at (axis cs:2,3) {2.68\%};
\node[font=\scriptsize, text=black] at (axis cs:3,3) {2.68\%};
\node[font=\scriptsize, text=black] at (axis cs:0,2) {5.47\%};
\node[font=\scriptsize, text=black] at (axis cs:1,2) {7.40\%};
\node[font=\scriptsize, text=black] at (axis cs:2,2) {9.54\%};
\node[font=\scriptsize, text=black] at (axis cs:3,2) {7.93\%};
\node[font=\scriptsize, text=black] at (axis cs:0,1) {7.07\%};
\node[font=\scriptsize, text=black] at (axis cs:1,1) {9.97\%};
\node[font=\scriptsize, text=black] at (axis cs:2,1) {7.93\%};
\node[font=\scriptsize, text=black] at (axis cs:3,1) {6.97\%};
\node[font=\scriptsize, text=white] at (axis cs:0,0) {27.12\%};
\node[font=\scriptsize, text=white] at (axis cs:1,0) {30.76\%};
\node[font=\scriptsize, text=white] at (axis cs:2,0) {64.52\%};
\node[font=\scriptsize, text=white] at (axis cs:3,0) {70.95\%};

\end{axis}
\end{tikzpicture}
\caption{Fall rate (\% of $n=933$ episodes) across four humanoid
embodiments and four VLN models.
Heatmap confirming DualVLN is the most stable model
across all embodiments.}
\label{fig:fall_rate}
\end{figure}

Fig.~\ref{fig:fall_rate} reveals substantial model--embodiment differences under physics-based execution. Fall rates remain low across Internal-A, G1, and Internal-B (2.7--10.0\%), but increase markedly on Unitree H1, reaching 64.5\% and 71.0\% for StreamVLN and NaVILA. These differences may reflect the combined effects of morphology, locomotion policies, and controller settings. Averaged across the four embodiments, DualVLN achieves the lowest Fall Rate, potentially benefiting from its continuous action interface. These results indicate that SR alone does not fully capture physical navigation stability.

\begin{figure*}
    \centering
    \includegraphics[width=0.72\linewidth]{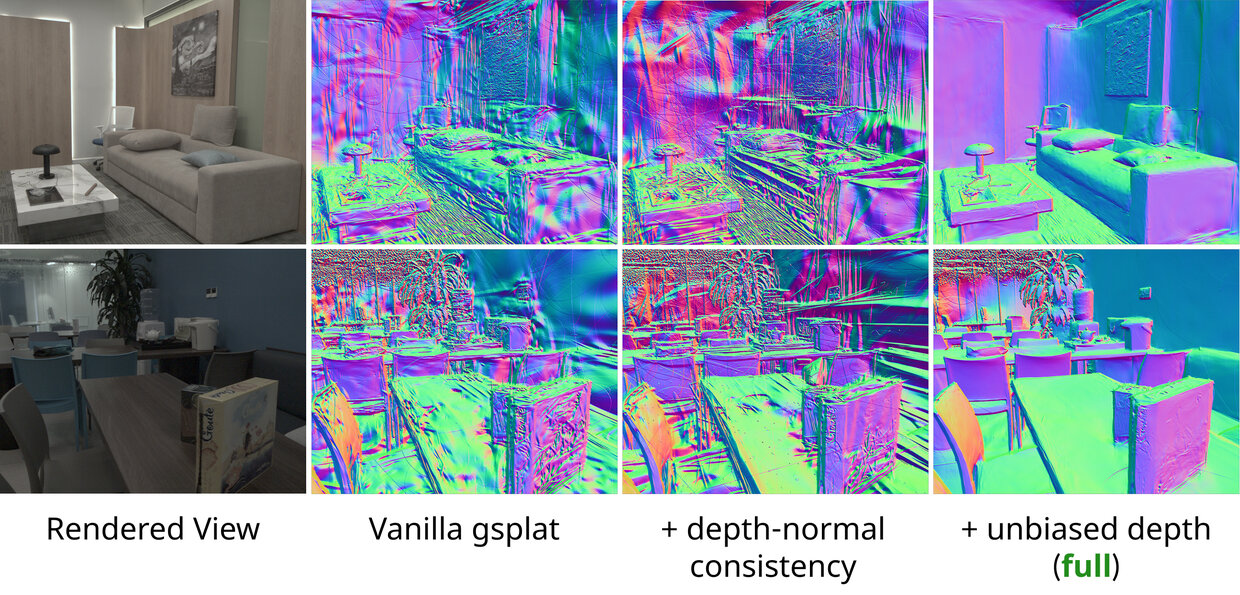}
    \caption{Qualitative geometry ablation of our 3DGS reconstruction pipeline. We visualize surface normals for two indoor scenes. Vanilla gsplat produces noisy, discontinuous normals. Adding depth-normal consistency improves local smoothness, while our full pipeline with unbiased depth yields the most coherent geometry.}
    \label{fig:rec_quality}
\end{figure*}

\subsection{Q3 --- 3DGS reconstruction quality}
\label{subsec:rec_quality} 

Figure \ref{fig:rec_quality} presents a qualitative ablation of our 3DGS reconstruction pipeline. Vanilla \texttt{gsplat}~\cite{ye2025gsplat} produces noisy and inconsistent surface normals, particularly around textureless walls, furniture boundaries, and thin structures, leading to collision meshes unsuitable for reliable physics simulation. Enforcing depth--normal consistency~\cite{Huang2DGS2024} improves local surface coherence but still leaves noticeable artifacts in large planar regions. Our full pipeline, combining unbiased depth rendering~\cite{Chen2025pgsr} with depth--normal consistency, produces substantially smoother and more geometrically consistent surfaces while preserving scene details. These improvements translate directly to higher-quality TSDF fusion results and more reliable collision meshes for Isaac Sim, enabling the use of reconstructed environments without requiring artist-authored assets.

\subsection{Q4 --- Real-World Correlation}
\label{subsec:real2sim}

A benchmark built on reconstructed scenes is useful only if agent
behavior remains consistent with the corresponding real environment.
We therefore conduct a pilot-scale sim-to-real study using paired
episode and trajectory analysis.

\noindent\textbf{Setup.}
We evaluate DualVLN in two environments, a \textbf{pantry} and a \textbf{studio room},
with 10 episodes per scene. Both scenes are captured and reconstructed
using a Unitree G1 robot equipped with an Intel RealSense D435i RGB-D
camera. The same DualVLN checkpoint is evaluated in simulation and in
the real environment, yielding $N=20$ paired episodes. An NVIDIA
Jetson AGX Orin handles onboard sensing and robot control, while the
inference server is hosted on an NVIDIA RTX A6000 Pro GPU. Real
trajectories are recorded using onboard localization and aligned
with the reconstructed scene coordinate frame.

\begin{figure}[t]
\centering
\begin{tikzpicture}
\begin{axis}[
  width=0.58\linewidth, height=4.6cm,
  xlabel={NE in sim (m)}, ylabel={NE in real (m)},
  xmin=0, xmax=10, ymin=0, ymax=10,
  tick label style={font=\scriptsize},
  label style={font=\scriptsize},
  title style={font=\scriptsize}, title={(a) Per-episode NE},
  grid=both, grid style={gray!20}]
\addplot[dashed, gray, domain=0:10] {x};

\addplot[only marks, mark=*, mark size=1.8pt, blue!70!black]
coordinates {
  (5.788729,5.6122123)
  (2.438652,2.7488585)
  (1.832970,3.6231221)
  (2.032074,3.6823440)
  (9.596671,9.8371210)
  (7.658379,7.3356243)
  (3.966443,3.8821230)
  (8.81232155,8.6712465)
  (5.344629,5.2456130)
  (6.411190,4.6212550)
};

\addplot[only marks, mark=square*, mark size=1.8pt, orange!85!black]
coordinates {
  (8.004254311,7.549973291)
  (7.830257498,5.719167112)
  (7.358541054,7.291942582)
  (9.380714825,8.599739683)
  (7.920946486,7.857134554)
  (2.345550000,2.775783000)
  (2.867780000,2.544432200)
  (1.892312000,2.375755100)
  (6.879912000,7.991822000)
  (5.145560000,6.312331000)
};
\end{axis}
\end{tikzpicture}%
\hfill
\begin{tikzpicture}
\begin{axis}[
  width=0.40\linewidth, height=4.6cm,
  boxplot/draw direction=y,
  ylabel={sim--real nDTW}, ymin=0.3, ymax=1.0,
  xtick={1,2}, xticklabels={Studio, Pantry},
  tick label style={font=\scriptsize},
  label style={font=\scriptsize},
  title style={font=\scriptsize}, title={(b) Trajectory similarity},
  grid=both, grid style={gray!20}]

\addplot+[
  boxplot prepared={
    lower whisker=0.345612,
    lower quartile=0.757728,
    median=0.873483,
    upper quartile=0.941718,
    upper whisker=0.983120
  },
  fill=blue!15, draw=blue!70!black] coordinates {};

\addplot+[
  boxplot prepared={
    lower whisker=0.419727,
    lower quartile=0.652764,
    median=0.738572,
    upper quartile=0.929910,
    upper whisker=0.951212
  },
  fill=orange!20, draw=orange!85!black] coordinates {};
\end{axis}
\end{tikzpicture}

\caption{Real-world validation at the episode level
($N=20$ paired episodes over two scenes). (a) Per-episode
navigation error in simulation versus reality (dashed: $y=x$);
the two are strongly correlated ($r=0.935$). (b) Distribution
of nDTW between each paired simulated and real trajectory,
grouped by scene.}
\label{fig:sim2real}
\end{figure}
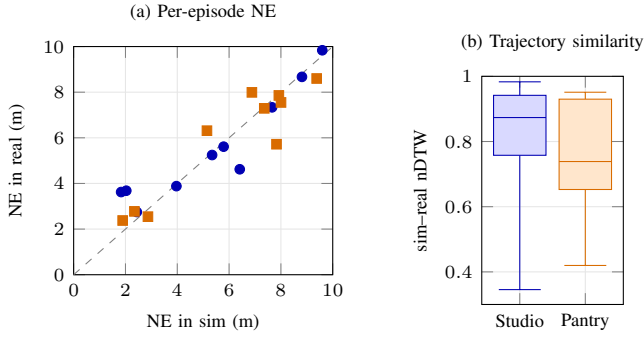

\noindent\textbf{Analysis.}
We evaluate sim-to-real consistency using endpoint error and trajectory
similarity. Per-episode navigation error is strongly correlated between
simulation and reality (Pearson $r=0.935$; Spearman $\rho=0.911$;
Fig.~\ref{fig:sim2real}a). The mean absolute NE difference is
$0.68$\,m, with a mean signed difference of only $0.04$\,m, indicating
little systematic endpoint bias. Paired trajectory similarity averages
$78.2 \pm 18.8$ nDTW, with scene-level means of 0.803 in the studio and
0.761 in the pantry (Fig.~\ref{fig:sim2real}b). These results indicate
that the reconstructed scenes largely preserve both navigation
difficulty and executed trajectory structure.

\noindent\textbf{Scope.}
This pilot study covers two scenes and one VLN checkpoint, so it does not
establish scene-level generalization. Instead, it provides initial
episode-level evidence that our reconstructions produce behavior
consistent with the corresponding real environments.

\subsection{Dataset Statistics}

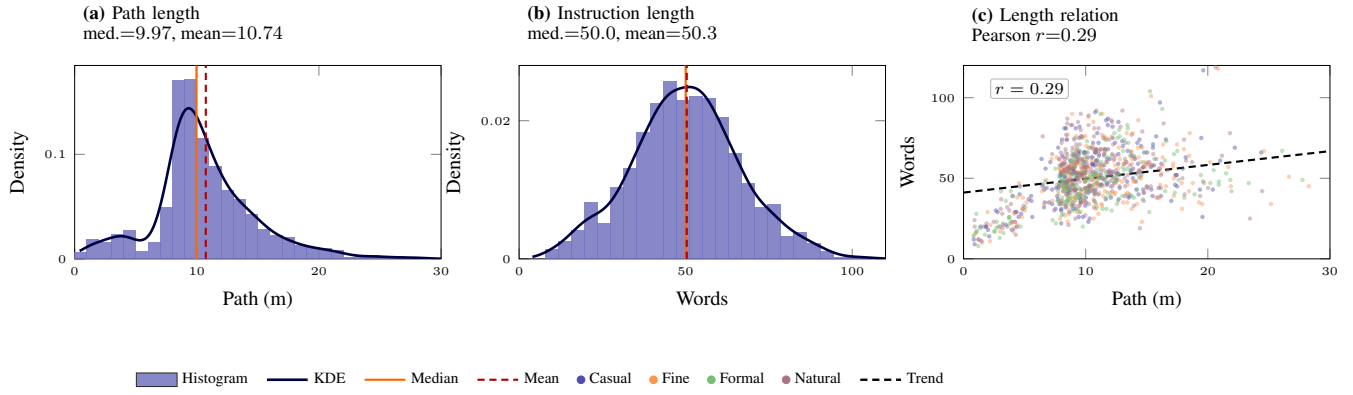
\begin{figure*}[t]
\centering
\resizebox{\textwidth}{!}{%
\begin{tikzpicture}
\begin{axis}[
    name=path,
    width=5.4cm, height=2.85cm, scale only axis,
    xmin=0, xmax=30, ymin=0, ymax=0.185,
    xtick={0,10,20,30}, ytick={0,0.1},
    scaled y ticks=false,
    yticklabel style={/pgf/number format/fixed,/pgf/number format/precision=1,
                      font=\fontsize{5.2}{5.8}\selectfont},
    x tick label style={font=\fontsize{5.2}{5.8}\selectfont},
    xlabel={Path (m)}, ylabel={Density},
    ylabel near ticks,
    xlabel style={font=\fontsize{8.5}{9.5}\selectfont},
    ylabel style={font=\fontsize{8.5}{9.5}\selectfont},
    title style={font=\fontsize{8.0}{9.0}\selectfont,align=left,
                 at={(0,1)},anchor=south west,yshift=0.5pt},
    title={\textbf{(a)} Path length\\[-0.5pt]
           {\fontsize{7.8}{8.8}\selectfont med.${=}9.97$, mean${=}10.74$}},
]
\fill[blue!55!black!55, opacity=0.85] (axis cs:0.00000,0) rectangle (axis cs:1.00000,0.00644468);
\fill[blue!55!black!55, opacity=0.85] (axis cs:1.00000,0) rectangle (axis cs:2.00000,0.01825994);
\fill[blue!55!black!55, opacity=0.85] (axis cs:2.00000,0) rectangle (axis cs:3.00000,0.01611171);
\fill[blue!55!black!55, opacity=0.85] (axis cs:3.00000,0) rectangle (axis cs:4.00000,0.02363050);
\fill[blue!55!black!55, opacity=0.85] (axis cs:4.00000,0) rectangle (axis cs:5.00000,0.02685285);
\fill[blue!55!black!55, opacity=0.85] (axis cs:5.00000,0) rectangle (axis cs:6.00000,0.00751880);
\fill[blue!55!black!55, opacity=0.85] (axis cs:6.00000,0) rectangle (axis cs:7.00000,0.01611171);
\fill[blue!55!black!55, opacity=0.85] (axis cs:7.00000,0) rectangle (axis cs:8.00000,0.04940924);
\fill[blue!55!black!55, opacity=0.85] (axis cs:8.00000,0) rectangle (axis cs:9.00000,0.17078410);
\fill[blue!55!black!55, opacity=0.85] (axis cs:9.00000,0) rectangle (axis cs:10.00000,0.17185822);
\fill[blue!55!black!55, opacity=0.85] (axis cs:10.00000,0) rectangle (axis cs:11.00000,0.11493018);
\fill[blue!55!black!55, opacity=0.85] (axis cs:11.00000,0) rectangle (axis cs:12.00000,0.08807734);
\fill[blue!55!black!55, opacity=0.85] (axis cs:12.00000,0) rectangle (axis cs:13.00000,0.06552095);
\fill[blue!55!black!55, opacity=0.85] (axis cs:13.00000,0) rectangle (axis cs:14.00000,0.05692803);
\fill[blue!55!black!55, opacity=0.85] (axis cs:14.00000,0) rectangle (axis cs:15.00000,0.04296455);
\fill[blue!55!black!55, opacity=0.85] (axis cs:15.00000,0) rectangle (axis cs:16.00000,0.02792696);
\fill[blue!55!black!55, opacity=0.85] (axis cs:16.00000,0) rectangle (axis cs:17.00000,0.02255639);
\fill[blue!55!black!55, opacity=0.85] (axis cs:17.00000,0) rectangle (axis cs:18.00000,0.02040816);
\fill[blue!55!black!55, opacity=0.85] (axis cs:18.00000,0) rectangle (axis cs:19.00000,0.01288937);
\fill[blue!55!black!55, opacity=0.85] (axis cs:19.00000,0) rectangle (axis cs:20.00000,0.01074114);
\fill[blue!55!black!55, opacity=0.85] (axis cs:20.00000,0) rectangle (axis cs:21.00000,0.00966702);
\fill[blue!55!black!55, opacity=0.85] (axis cs:21.00000,0) rectangle (axis cs:22.00000,0.00859291);
\fill[blue!55!black!55, opacity=0.85] (axis cs:22.00000,0) rectangle (axis cs:23.00000,0.00107411);
\fill[blue!55!black!55, opacity=0.85] (axis cs:23.00000,0) rectangle (axis cs:24.00000,0.00214823);
\fill[blue!55!black!55, opacity=0.85] (axis cs:24.00000,0) rectangle (axis cs:25.00000,0.00322234);
\fill[blue!55!black!55, opacity=0.85] (axis cs:25.00000,0) rectangle (axis cs:26.00000,0.00107411);
\fill[blue!55!black!55, opacity=0.85] (axis cs:26.00000,0) rectangle (axis cs:27.00000,0.00214823);
\fill[blue!55!black!55, opacity=0.85] (axis cs:27.00000,0) rectangle (axis cs:28.00000,0.00107411);
\fill[blue!55!black!55, opacity=0.85] (axis cs:28.00000,0) rectangle (axis cs:29.00000,0.00107411);
\addplot[color=blue!25!black, line width=1.1pt, forget plot] coordinates {
    (0.40815,0.00749555) (0.55685,0.00846443) (0.70556,0.00943514) (0.85426,0.01039041) (1.00296,0.01131536)
    (1.15166,0.01219866) (1.30037,0.01303324) (1.44907,0.01381660) (1.59777,0.01455057) (1.74647,0.01524049)
    (1.89518,0.01589418) (2.04388,0.01652040) (2.19258,0.01712742) (2.34129,0.01772152) (2.48999,0.01830576)
    (2.63869,0.01887912) (2.78739,0.01943600) (2.93610,0.01996633) (3.08480,0.02045606) (3.23350,0.02088815)
    (3.38221,0.02124395) (3.53091,0.02150487) (3.67961,0.02165427) (3.82831,0.02167934) (3.97702,0.02157298)
    (4.12572,0.02133544) (4.27442,0.02097561) (4.42312,0.02051191) (4.57183,0.01997264) (4.72053,0.01939594)
    (4.86923,0.01882913) (5.01794,0.01832777) (5.16664,0.01795440) (5.31534,0.01777703) (5.46404,0.01786760)
    (5.61275,0.01830032) (5.76145,0.01914993) (5.91015,0.02048981) (6.05886,0.02238980) (6.20756,0.02491359)
    (6.35626,0.02811562) (6.50496,0.03203746) (6.65367,0.03670376) (6.80237,0.04211805) (6.95107,0.04825879)
    (7.09977,0.05507609) (7.24848,0.06248976) (7.39718,0.07038907) (7.54588,0.07863466) (7.69459,0.08706271)
    (7.84329,0.09549132) (7.99199,0.10372861) (8.14069,0.11158213) (8.28940,0.11886862) (8.43810,0.12542332)
    (8.58680,0.13110813) (8.73550,0.13581781) (8.88421,0.13948389) (9.03291,0.14207597) (9.18161,0.14360052)
    (9.33032,0.14409759) (9.47902,0.14363562) (9.62772,0.14230520) (9.77642,0.14021224) (9.92513,0.13747114)
    (10.07383,0.13419854) (10.22253,0.13050794) (10.37124,0.12650550) (10.51994,0.12228706) (10.66864,0.11793647)
    (10.81734,0.11352500) (10.96605,0.10911164) (11.11475,0.10474416) (11.26345,0.10046043) (11.41215,0.09629009)
    (11.56086,0.09225604) (11.70956,0.08837580) (11.85826,0.08466262) (12.00697,0.08112621) (12.15567,0.07777318)
    (12.30437,0.07460722) (12.45307,0.07162909) (12.60178,0.06883642) (12.75048,0.06622364) (12.89918,0.06378182)
    (13.04788,0.06149875) (13.19659,0.05935918) (13.34529,0.05734535) (13.49399,0.05543764) (13.64270,0.05361554)
    (13.79140,0.05185870) (13.94010,0.05014801) (14.08880,0.04846666) (14.23751,0.04680103) (14.38621,0.04514138)
    (14.53491,0.04348214) (14.68362,0.04182190) (14.83232,0.04016315) (14.98102,0.03851157) (15.12972,0.03687529)
    (15.27843,0.03526393) (15.42713,0.03368770) (15.57583,0.03215655) (15.72453,0.03067947) (15.87324,0.02926399)
    (16.02194,0.02791581) (16.17064,0.02663866) (16.31935,0.02543427) (16.46805,0.02430240) (16.61675,0.02324102)
    (16.76545,0.02224655) (16.91416,0.02131415) (17.06286,0.02043811) (17.21156,0.01961228) (17.36027,0.01883051)
    (17.50897,0.01808717) (17.65767,0.01737754) (17.80637,0.01669811) (17.95508,0.01604676) (18.10378,0.01542272)
    (18.25248,0.01482633) (18.40118,0.01425873) (18.54989,0.01372135) (18.69859,0.01321543) (18.84729,0.01274163)
    (18.99600,0.01229969) (19.14470,0.01188828) (19.29340,0.01150502) (19.44210,0.01114662) (19.59081,0.01080914)
    (19.73951,0.01048828) (19.88821,0.01017968) (20.03691,0.00987909) (20.18562,0.00958257) (20.33432,0.00928655)
    (20.48302,0.00898783) (20.63173,0.00868356) (20.78043,0.00837129) (20.92913,0.00804899) (21.07783,0.00771516)
    (21.22654,0.00736896) (21.37524,0.00701046) (21.52394,0.00664074) (21.67265,0.00626210) (21.82135,0.00587803)
    (21.97005,0.00549319) (22.11875,0.00511318) (22.26746,0.00474417) (22.41616,0.00439252) (22.56486,0.00406428)
    (22.71356,0.00376467) (22.86227,0.00349767) (23.01097,0.00326570) (23.15967,0.00306943) (23.30838,0.00290773)
    (23.45708,0.00277786) (23.60578,0.00267567) (23.75448,0.00259604) (23.90319,0.00253332) (24.05189,0.00248176)
    (24.20059,0.00243600) (24.34929,0.00239139) (24.49800,0.00234429) (24.64670,0.00229222) (24.79540,0.00223388)
    (24.94411,0.00216910) (25.09281,0.00209864) (25.24151,0.00202393) (25.39021,0.00194687) (25.53892,0.00186950)
    (25.68762,0.00179379) (25.83632,0.00172146) (25.98503,0.00165382) (26.13373,0.00159174) (26.28243,0.00153562)
    (26.43113,0.00148542) (26.57984,0.00144068) (26.72854,0.00140060) (26.87724,0.00136411) (27.02594,0.00132990)
    (27.17465,0.00129649) (27.32335,0.00126235) (27.47205,0.00122592) (27.62076,0.00118575) (27.76946,0.00114062)
    (27.91816,0.00108960) (28.06686,0.00103220) (28.21557,0.00096839) (28.36427,0.00089866) (28.51297,0.00082397)
    (28.66168,0.00074570) (28.81038,0.00066554) (28.95908,0.00058534) (29.10778,0.00050696) (29.25649,0.00043214)
    (29.40519,0.00036237) (29.55389,0.00029879) (29.70259,0.00024218) (29.85130,0.00019288) (30.00000,0.00015093)
};
\draw[orange!80!red, line width=0.9pt] (axis cs:9.9735,0) -- (axis cs:9.9735,0.185);
\draw[red!70!black, densely dashed, line width=0.9pt] (axis cs:10.7374,0) -- (axis cs:10.7374,0.185);
\end{axis}

\begin{axis}[
    name=words,
    at={(path.south east)}, anchor=south west, xshift=1.15cm,
    width=5.4cm, height=2.85cm, scale only axis,
    xmin=0, xmax=110, ymin=0, ymax=0.028,
    xtick={0,50,100}, ytick={0,0.02},
    scaled y ticks=false,
    yticklabel style={/pgf/number format/fixed,/pgf/number format/precision=2,
                      font=\fontsize{5.2}{5.8}\selectfont},
    x tick label style={font=\fontsize{5.2}{5.8}\selectfont},
    xlabel={Words}, ylabel={Density},
    ylabel near ticks,
    xlabel style={font=\fontsize{8.5}{9.5}\selectfont},
    ylabel style={font=\fontsize{8.5}{9.5}\selectfont},
    title style={font=\fontsize{8.0}{9.0}\selectfont,align=left,
                 at={(0,1)},anchor=south west,yshift=0.5pt},
    title={\textbf{(b)} Instruction length\\[-0.5pt]
           {\fontsize{7.8}{8.8}\selectfont med.${=}50.0$, mean${=}50.3$}},
]
\fill[blue!55!black!55, opacity=0.85] (axis cs:7.85714,0) rectangle (axis cs:11.78571,0.00136852);
\fill[blue!55!black!55, opacity=0.85] (axis cs:11.78571,0) rectangle (axis cs:15.71429,0.00246334);
\fill[blue!55!black!55, opacity=0.85] (axis cs:15.71429,0) rectangle (axis cs:19.64286,0.00410557);
\fill[blue!55!black!55, opacity=0.85] (axis cs:19.64286,0) rectangle (axis cs:23.57143,0.00821114);
\fill[blue!55!black!55, opacity=0.85] (axis cs:23.57143,0) rectangle (axis cs:27.50000,0.00520039);
\fill[blue!55!black!55, opacity=0.85] (axis cs:27.50000,0) rectangle (axis cs:31.42857,0.01040078);
\fill[blue!55!black!55, opacity=0.85] (axis cs:31.42857,0) rectangle (axis cs:35.35714,0.01313783);
\fill[blue!55!black!55, opacity=0.85] (axis cs:35.35714,0) rectangle (axis cs:39.28571,0.01833822);
\fill[blue!55!black!55, opacity=0.85] (axis cs:39.28571,0) rectangle (axis cs:43.21429,0.02244379);
\fill[blue!55!black!55, opacity=0.85] (axis cs:43.21429,0) rectangle (axis cs:47.14286,0.02572825);
\fill[blue!55!black!55, opacity=0.85] (axis cs:47.14286,0) rectangle (axis cs:51.07143,0.02299120);
\fill[blue!55!black!55, opacity=0.85] (axis cs:51.07143,0) rectangle (axis cs:55.00000,0.02353861);
\fill[blue!55!black!55, opacity=0.85] (axis cs:55.00000,0) rectangle (axis cs:58.92857,0.02326491);
\fill[blue!55!black!55, opacity=0.85] (axis cs:58.92857,0) rectangle (axis cs:62.85714,0.02052786);
\fill[blue!55!black!55, opacity=0.85] (axis cs:62.85714,0) rectangle (axis cs:66.78571,0.01532747);
\fill[blue!55!black!55, opacity=0.85] (axis cs:66.78571,0) rectangle (axis cs:70.71429,0.01094819);
\fill[blue!55!black!55, opacity=0.85] (axis cs:70.71429,0) rectangle (axis cs:74.64286,0.00739003);
\fill[blue!55!black!55, opacity=0.85] (axis cs:74.64286,0) rectangle (axis cs:78.57143,0.00793744);
\fill[blue!55!black!55, opacity=0.85] (axis cs:78.57143,0) rectangle (axis cs:82.50000,0.00328446);
\fill[blue!55!black!55, opacity=0.85] (axis cs:82.50000,0) rectangle (axis cs:86.42857,0.00383187);
\fill[blue!55!black!55, opacity=0.85] (axis cs:86.42857,0) rectangle (axis cs:90.35714,0.00218964);
\fill[blue!55!black!55, opacity=0.85] (axis cs:90.35714,0) rectangle (axis cs:94.28571,0.00109482);
\fill[blue!55!black!55, opacity=0.85] (axis cs:94.28571,0) rectangle (axis cs:98.21429,0.00027370);
\fill[blue!55!black!55, opacity=0.85] (axis cs:98.21429,0) rectangle (axis cs:102.14286,0.00027370);
\fill[blue!55!black!55, opacity=0.85] (axis cs:102.14286,0) rectangle (axis cs:106.07143,0.00027370);
\addplot[color=blue!25!black, line width=1.1pt, forget plot] coordinates {
    (4.00000,0.00028891) (4.53266,0.00035200) (5.06533,0.00042408) (5.59799,0.00050547) (6.13065,0.00059630)
    (6.66332,0.00069659) (7.19598,0.00080624) (7.72864,0.00092510) (8.26131,0.00105293) (8.79397,0.00118953)
    (9.32663,0.00133474) (9.85930,0.00148847) (10.39196,0.00165075) (10.92462,0.00182168) (11.45729,0.00200146)
    (11.98995,0.00219028) (12.52261,0.00238832) (13.05528,0.00259560) (13.58794,0.00281192) (14.12060,0.00303675)
    (14.65327,0.00326919) (15.18593,0.00350787) (15.71859,0.00375097) (16.25126,0.00399627) (16.78392,0.00424120)
    (17.31658,0.00448301) (17.84925,0.00471893) (18.38191,0.00494636) (18.91457,0.00516313) (19.44724,0.00536763)
    (19.97990,0.00555907) (20.51256,0.00573756) (21.04523,0.00590417) (21.57789,0.00606099) (22.11055,0.00621100)
    (22.64322,0.00635796) (23.17588,0.00650621) (23.70854,0.00666046) (24.24121,0.00682550) (24.77387,0.00700602)
    (25.30653,0.00720630) (25.83920,0.00743013) (26.37186,0.00768055) (26.90452,0.00795987) (27.43719,0.00826953)
    (27.96985,0.00861018) (28.50251,0.00898168) (29.03518,0.00938320) (29.56784,0.00981327) (30.10050,0.01026989)
    (30.63317,0.01075064) (31.16583,0.01125277) (31.69849,0.01177334) (32.23116,0.01230924) (32.76382,0.01285738)
    (33.29648,0.01341473) (33.82915,0.01397845) (34.36181,0.01454594) (34.89447,0.01511488) (35.42714,0.01568332)
    (35.95980,0.01624961) (36.49246,0.01681236) (37.02513,0.01737038) (37.55779,0.01792253) (38.09045,0.01846759)
    (38.62312,0.01900410) (39.15578,0.01953027) (39.68844,0.02004386) (40.22111,0.02054218) (40.75377,0.02102213)
    (41.28643,0.02148029) (41.81910,0.02191317) (42.35176,0.02231742) (42.88442,0.02269008) (43.41709,0.02302892)
    (43.94975,0.02333257) (44.48241,0.02360075) (45.01508,0.02383431) (45.54774,0.02403513) (46.08040,0.02420601)
    (46.61307,0.02435033) (47.14573,0.02447176) (47.67839,0.02457384) (48.21106,0.02465961) (48.74372,0.02473125)
    (49.27638,0.02478990) (49.80905,0.02483542) (50.34171,0.02486648) (50.87437,0.02488061) (51.40704,0.02487444)
    (51.93970,0.02484399) (52.47236,0.02478506) (53.00503,0.02469355) (53.53769,0.02456581) (54.07035,0.02439892)
    (54.60302,0.02419093) (55.13568,0.02394091) (55.66834,0.02364906) (56.20101,0.02331662) (56.73367,0.02294579)
    (57.26633,0.02253957) (57.79899,0.02210156) (58.33166,0.02163579) (58.86432,0.02114649) (59.39698,0.02063790)
    (59.92965,0.02011409) (60.46231,0.01957880) (60.99497,0.01903534) (61.52764,0.01848645) (62.06030,0.01793433)
    (62.59296,0.01738059) (63.12563,0.01682637) (63.65829,0.01627242) (64.19095,0.01571926) (64.72362,0.01516741)
    (65.25628,0.01461752) (65.78894,0.01407061) (66.32161,0.01352817) (66.85427,0.01299224) (67.38693,0.01246541)
    (67.91960,0.01195070) (68.45226,0.01145140) (68.98492,0.01097084) (69.51759,0.01051206) (70.05025,0.01007754)
    (70.58291,0.00966894) (71.11558,0.00928690) (71.64824,0.00893095) (72.18090,0.00859951) (72.71357,0.00829005)
    (73.24623,0.00799927) (73.77889,0.00772345) (74.31156,0.00745878) (74.84422,0.00720172) (75.37688,0.00694925)
    (75.90955,0.00669917) (76.44221,0.00645016) (76.97487,0.00620182) (77.50754,0.00595461) (78.04020,0.00570961)
    (78.57286,0.00546836) (79.10553,0.00523257) (79.63819,0.00500388) (80.17085,0.00478367) (80.70352,0.00457290)
    (81.23618,0.00437202) (81.76884,0.00418101) (82.30151,0.00399939) (82.83417,0.00382634) (83.36683,0.00366081)
    (83.89950,0.00350167) (84.43216,0.00334777) (84.96482,0.00319809) (85.49749,0.00305175) (86.03015,0.00290806)
    (86.56281,0.00276649) (87.09548,0.00262669) (87.62814,0.00248843) (88.16080,0.00235156) (88.69347,0.00221606)
    (89.22613,0.00208195) (89.75879,0.00194933) (90.29146,0.00181842) (90.82412,0.00168953) (91.35678,0.00156310)
    (91.88945,0.00143968) (92.42211,0.00131997) (92.95477,0.00120471) (93.48744,0.00109474) (94.02010,0.00099087)
    (94.55276,0.00089388) (95.08543,0.00080440) (95.61809,0.00072293) (96.15075,0.00064975) (96.68342,0.00058492)
    (97.21608,0.00052827) (97.74874,0.00047939) (98.28141,0.00043769) (98.81407,0.00040240) (99.34673,0.00037264)
    (99.87940,0.00034749) (100.41206,0.00032597) (100.94472,0.00030717) (101.47739,0.00029023) (102.01005,0.00027442)
    (102.54271,0.00025912) (103.07538,0.00024386) (103.60804,0.00022834) (104.14070,0.00021239) (104.67337,0.00019597)
    (105.20603,0.00017917) (105.73869,0.00016215) (106.27136,0.00014513) (106.80402,0.00012839) (107.33668,0.00011219)
    (107.86935,0.00009679) (108.40201,0.00008240) (108.93467,0.00006921) (109.46734,0.00005733) (110.00000,0.00004682)
};
\draw[orange!80!red, line width=0.9pt] (axis cs:50.0,0) -- (axis cs:50.0,0.028);
\draw[red!70!black, densely dashed, line width=0.9pt] (axis cs:50.3355,0) -- (axis cs:50.3355,0.028);
\end{axis}

\begin{axis}[
    name=scat,
    at={(words.south east)}, anchor=south west, xshift=1.15cm,
    width=5.4cm, height=2.85cm, scale only axis,
    xmin=0, xmax=30, ymin=0, ymax=120,
    xtick={0,10,20,30}, ytick={0,50,100},
    xlabel={Path (m)}, ylabel={Words},
    ylabel near ticks,
    xlabel style={font=\fontsize{8.5}{9.5}\selectfont},
    ylabel style={font=\fontsize{8.5}{9.5}\selectfont},
    tick label style={font=\fontsize{5.2}{5.8}\selectfont},
    title style={font=\fontsize{8.0}{9.0}\selectfont,align=left,
                 at={(0,1)},anchor=south west,yshift=0.5pt},
    title={\textbf{(c)} Length relation\\[-0.5pt]
           {\fontsize{7.8}{8.8}\selectfont Pearson $r{=}0.29$}},
]
\addplot[only marks, mark=*, mark size=0.7pt, blue!55!black!70, opacity=0.45, forget plot] coordinates {
    (7.8421,34) (8.3023,51) (8.6466,41) (24.3761,66) (9.9735,33) (16.1613,39) (9.4256,26) (8.1534,42)
    (13.5890,39) (21.0126,61) (11.0480,40) (9.6121,35) (10.0150,50) (14.1609,53) (12.0575,38) (17.9464,56)
    (14.3277,40) (12.5653,49) (4.6201,45) (10.8516,61) (14.1516,43) (9.5031,82) (17.3218,46) (8.4614,47)
    (17.3453,52) (17.4371,52) (15.3811,43) (17.4916,45) (12.6612,53) (19.3534,43) (15.5014,69) (7.2436,43)
    (17.0114,36) (20.4091,43) (16.1986,59) (8.9320,56) (10.7881,58) (11.2616,52) (12.3301,47) (7.8103,62)
    (9.5476,53) (10.9485,78) (7.7851,44) (15.4039,58) (15.0107,45) (9.9383,54) (10.0505,35) (14.4040,74)
    (10.2352,41) (9.7757,26) (13.5143,38) (10.2224,23) (10.0299,38) (7.9727,48) (10.6170,58) (8.9467,31)
    (9.6925,53) (9.1929,31) (19.5620,26) (15.9842,44) (10.3182,25) (9.4587,38) (14.0327,62) (11.1083,51)
    (12.1962,82) (9.9396,48) (22.2076,69) (13.4366,67) (14.5401,70) (14.7879,66) (9.8686,60) (9.0540,60)
    (9.0306,65) (12.6722,59) (16.0828,97) (19.6260,117) (8.2614,81) (9.2307,43) (14.7089,43) (9.7528,62)
    (12.6832,85) (14.2792,59) (9.2958,50) (7.5801,44) (12.9945,73) (7.6656,39) (4.1952,29) (14.7262,57)
    (8.5290,36) (10.1521,63) (1.3954,20) (7.2590,34) (10.4042,45) (4.3345,29) (6.7152,56) (7.7495,69)
    (6.5900,32) (11.3262,46) (3.9467,18) (6.9909,35) (4.0105,39) (6.6516,35) (10.6166,54) (2.1029,14)
    (6.3824,62) (9.3074,60) (13.8778,56) (11.8327,87) (11.0907,64) (7.7393,37) (11.7437,62) (13.3862,48)
    (15.5143,71) (4.3777,20) (5.3332,37) (1.4626,11) (5.6278,26) (9.9945,55) (4.8634,31) (8.8954,46)
    (4.5085,26) (14.1099,48) (8.2643,51) (9.8669,36) (0.9071,9) (8.7557,43) (7.9852,52) (0.8392,12)
    (8.3683,47) (14.2326,72) (10.7826,91) (9.9674,66) (9.1173,42) (8.9593,56) (8.0072,58) (10.8184,66)
    (10.7116,36) (11.8274,68) (9.2363,64) (2.0082,18) (1.8278,20) (9.1588,49) (2.6388,30) (10.1634,43)
    (9.2723,62) (8.8119,53) (16.0706,56) (11.2336,46) (8.9311,56) (9.3076,52) (10.7917,41) (13.8625,50)
    (7.8729,51) (12.2303,54) (12.7861,44) (2.7892,29) (11.6736,86) (1.8595,20) (9.3215,40) (9.8920,66)
    (8.4655,54) (9.4566,65) (9.0519,43) (8.1848,59) (11.5154,65) (7.1042,60) (6.4796,46) (11.8059,83)
    (8.0996,47) (13.7614,64) (11.9700,62) (9.1801,81) (11.8089,54) (9.4960,67) (9.8407,66) (6.6040,43)
    (8.3770,47) (9.1053,57) (11.5537,68) (11.0493,72) (8.3972,81) (10.4774,63) (10.2805,45) (8.0838,64)
    (4.7192,28) (15.6003,58) (12.6817,65) (10.7251,52) (11.9886,52) (5.5343,39) (14.3798,54) (8.8660,49)
    (12.5710,53) (9.5783,35) (3.4718,21) (12.4024,72) (8.6235,56) (9.9472,57) (8.0643,53) (8.5815,42)
    (8.4720,74) (10.8754,68) (12.6996,37) (10.8118,53) (7.9744,53) (8.0050,31) (11.5303,49) (11.7554,47)
    (16.5684,74) (18.1853,63) (15.7777,76) (4.5925,40) (9.2223,50) (9.4574,75) (10.5257,59) (10.5657,59)
    (10.8083,43) (13.7066,54) (10.7592,66) (11.7292,46) (8.1633,54) (8.6762,29) (11.6420,37) (8.3617,32)
    (8.2005,54) (7.9956,22) (8.1824,45) (9.1191,54) (5.5556,36) (7.5249,31) (10.5993,72) (8.9968,54)
    (7.9916,30) (10.9640,61) (13.3367,60) (8.6776,56) (9.8669,53) (10.5430,42) (9.3769,66) (12.2076,75)
    (15.3265,57) (15.4704,39) (10.1936,44) (19.4661,75) (13.0446,70) (3.9664,23) (9.2386,53) (11.2428,68)
    (8.7219,61) (8.0636,74)
};
\addplot[only marks, mark=*, mark size=0.7pt, orange!80!red!70, opacity=0.45, forget plot] coordinates {
    (9.9436,49) (14.1911,34) (13.4097,35) (9.6148,51) (13.4362,41) (8.2341,48) (10.3076,51) (12.4319,47)
    (10.8774,34) (25.1563,34) (8.6129,28) (13.4495,46) (11.0476,28) (21.6657,57) (17.1461,57) (8.1614,44)
    (8.1201,36) (12.9915,48) (8.6964,53) (9.0695,45) (16.5564,45) (8.5290,23) (18.5106,78) (10.2192,69)
    (12.3431,40) (8.8850,90) (18.4793,64) (11.6927,74) (10.5237,45) (17.1571,59) (11.5767,42) (19.5465,34)
    (9.6267,57) (10.8153,31) (10.7335,40) (13.9107,48) (28.2700,45) (26.6475,51) (18.7664,47) (8.7263,39)
    (24.5806,67) (14.3983,50) (9.9777,76) (8.1280,54) (8.9012,48) (10.0638,44) (10.8923,52) (9.1214,57)
    (16.2039,53) (14.9075,59) (8.8231,37) (18.4510,42) (17.2282,50) (12.0312,41) (11.2978,34) (13.7634,31)
    (12.6833,56) (13.6995,67) (9.8611,35) (10.6466,25) (13.9890,48) (7.9052,25) (8.2526,53) (8.1219,34)
    (10.6845,46) (10.0602,70) (14.6049,23) (17.8867,31) (10.6369,37) (15.9720,38) (17.8641,47) (17.2050,39)
    (21.8939,45) (19.2825,35) (13.0733,35) (14.5362,60) (8.4831,48) (20.1843,60) (13.1298,59) (14.3480,54)
    (16.3276,58) (8.6467,51) (15.3556,101) (9.6313,42) (16.0028,38) (11.3204,77) (16.0336,56) (10.0471,42)
    (10.7600,50) (13.7876,59) (9.3039,40) (9.1927,37) (8.8463,57) (11.6718,75) (10.0390,63) (9.2885,29)
    (11.3664,90) (8.4422,45) (16.2993,63) (16.0937,46) (20.8493,118) (14.9590,44) (9.5430,40) (8.7543,36)
    (11.0528,52) (7.7825,40) (7.2069,62) (6.4143,42) (8.5144,77) (8.7486,50) (1.5272,10) (8.2179,57)
    (9.4383,59) (6.7611,49) (9.2313,76) (9.9641,44) (11.5088,76) (8.8053,56) (9.1970,38) (4.1413,28)
    (4.7789,26) (11.0275,46) (0.8729,15) (4.8506,38) (9.6429,62) (11.4311,45) (6.6057,34) (4.8166,31)
    (13.8604,78) (8.5802,31) (2.1688,11) (4.0698,17) (8.7059,58) (7.9413,56) (13.3562,78) (10.0734,55)
    (8.4428,56) (10.6337,85) (8.3459,45) (14.5950,69) (15.3396,58) (14.3264,63) (11.3341,33) (9.8081,76)
    (1.0560,17) (10.2693,58) (8.4713,56) (9.2980,59) (10.2140,88) (11.5481,66) (11.1021,41) (12.6919,49)
    (9.8492,53) (5.0109,30) (15.0134,60) (7.9144,61) (7.6136,63) (12.4189,58) (7.8467,56) (9.8812,35)
    (1.9658,29) (8.9274,59) (2.2047,21) (10.1620,57) (12.2992,61) (9.0394,61) (4.5378,47) (9.2998,65)
    (5.1664,35) (10.9762,45) (9.3457,49) (7.9816,54) (9.3333,54) (11.0443,50) (14.2850,69) (14.5523,50)
    (9.0364,57) (9.9919,48) (11.0203,58) (11.4255,39) (16.3461,71) (9.4362,33) (11.5275,68) (12.7869,62)
    (18.0343,68) (9.6417,54) (17.4828,71) (14.3039,86) (10.5856,70) (9.8723,56) (12.7069,69) (13.7841,62)
    (11.2511,56) (4.4002,38) (4.4078,21) (8.0058,45) (12.4284,41) (10.4855,61) (9.0999,39) (13.2930,40)
    (9.8226,56) (10.8671,47) (9.8649,50) (10.5806,53) (10.4217,49) (7.2636,36) (8.7606,54) (9.5913,37)
    (9.0597,43) (10.5233,43) (8.1586,42) (12.1380,76) (2.9857,17) (9.4681,29) (11.2092,68) (13.2952,60)
    (9.9124,53) (8.3157,27) (14.8858,57) (4.2068,23) (8.2046,55) (9.2932,68) (3.9157,22) (9.3961,60)
};
\addplot[only marks, mark=*, mark size=0.7pt, green!55!black!55, opacity=0.45, forget plot] coordinates {
    (9.9086,43) (9.7614,20) (11.7023,31) (8.8113,45) (12.9972,39) (12.0193,39) (10.1138,42) (20.4763,55)
    (8.3786,34) (12.7668,52) (21.8449,49) (13.9406,38) (7.2864,24) (13.6018,40) (8.0431,49) (20.8912,46)
    (8.0411,27) (13.3646,56) (17.5044,67) (12.3976,48) (9.9283,38) (17.6089,49) (19.6962,47) (10.7721,35)
    (14.2217,43) (9.3933,40) (9.3622,50) (8.9769,65) (11.2938,60) (10.5023,51) (8.3485,66) (8.1300,28)
    (15.2072,29) (10.0300,44) (8.2870,42) (9.8900,65) (20.8961,48) (26.0815,67) (18.9197,40) (3.8702,42)
    (27.7455,49) (17.2930,41) (23.3405,47) (8.4953,43) (12.1253,30) (9.9662,54) (10.3303,69) (11.1967,90)
    (9.5136,53) (13.7556,39) (9.9678,57) (10.9375,53) (8.8308,41) (18.0697,38) (10.1817,50) (9.5112,37)
    (8.4196,42) (9.0263,62) (8.9914,30) (8.7082,48) (10.6235,52) (8.5356,38) (10.7399,41) (8.2575,44)
    (11.6559,50) (8.2564,35) (10.8730,55) (12.6953,52) (9.5706,29) (9.5390,43) (10.8819,67) (12.7350,80)
    (11.3828,31) (8.1813,45) (9.9036,60) (21.9407,42) (20.2035,47) (8.1468,48) (10.9292,32) (11.0650,50)
    (16.5099,42) (9.3616,31) (19.4323,52) (10.2881,43) (12.6395,36) (21.4994,53) (17.2260,65) (7.6941,36)
    (16.2741,93) (9.2866,53) (12.7987,76) (10.0966,73) (12.5985,75) (15.2639,104) (16.6063,73) (13.0268,57)
    (13.4107,83) (11.6782,38) (15.0123,66) (8.0837,38) (8.4634,43) (8.2029,45) (8.4298,53) (7.8551,52)
    (1.0104,16) (6.6772,48) (3.2513,22) (10.2180,49) (7.7460,42) (1.0700,17) (8.5154,47) (3.7365,26)
    (9.8569,52) (9.7222,59) (1.7485,15) (11.1451,56) (10.3565,52) (13.0562,67) (11.6557,53) (13.4905,72)
    (13.9412,64) (11.6989,47) (8.8999,34) (2.7232,23) (2.9649,13) (2.6812,19) (7.5617,22) (8.2166,52)
    (3.7364,26) (14.7399,60) (10.0818,62) (11.2269,50) (8.5190,46) (8.7837,56) (2.8391,20) (11.8176,63)
    (11.5306,58) (9.1468,43) (10.6311,46) (8.6259,44) (10.6652,52) (8.8171,38) (11.5932,74) (9.5070,40)
    (12.1844,64) (8.7548,61) (8.9123,53) (7.8889,47) (11.5933,67) (13.6251,64) (13.1539,74) (3.0590,42)
    (11.1959,45) (2.5050,17) (8.2794,44) (9.0581,58) (1.0526,28) (9.9248,48) (0.8163,14) (1.3313,19)
    (12.3600,78) (12.5191,48) (10.9078,48) (8.8589,52) (8.5973,37) (9.1510,64) (7.6674,46) (9.3054,69)
    (7.7802,36) (9.1457,36) (7.8403,58) (9.3052,51) (8.3533,45) (4.6626,20) (9.5842,37) (5.0661,32)
    (10.8554,61) (8.1303,49) (8.8580,61) (1.2135,8) (9.0781,52) (8.8821,67) (4.0564,39) (3.7275,25)
    (8.8706,33) (8.3263,48) (9.8901,40) (14.4840,54) (12.4549,52) (15.6265,62) (3.4567,36) (9.5630,40)
    (3.2680,16) (2.5596,17) (9.1114,38) (9.4698,44) (16.5106,61) (9.6766,53) (11.7958,41) (11.7189,51)
    (14.6631,37) (12.1496,49) (9.1214,47) (13.6225,55) (10.3199,45) (6.7175,44) (8.5142,44) (9.5324,38)
    (12.5669,34) (8.4437,36) (7.9995,35) (8.5492,39) (14.3922,51) (12.2981,50) (9.7065,75) (7.5303,56)
    (8.9541,47) (9.4189,62) (9.0018,45) (9.0923,75) (9.1070,57) (13.4445,81) (11.1457,51) (9.9347,66)
    (13.1326,55) (3.7844,22) (8.5894,34)
};
\addplot[only marks, mark=*, mark size=0.7pt, purple!55!black!55, opacity=0.45, forget plot] coordinates {
    (9.1465,43) (9.1569,25) (13.2285,45) (13.4410,52) (13.9448,33) (18.0553,74) (15.2191,60) (17.9161,50)
    (16.1620,52) (8.4037,40) (8.5800,57) (8.2530,55) (21.1644,64) (18.2260,61) (14.5633,52) (9.9113,57)
    (9.6642,51) (9.2466,50) (9.7089,53) (11.8919,52) (19.3747,39) (21.8132,78) (14.3495,55) (13.6132,58)
    (8.4506,40) (8.9361,69) (12.4394,64) (14.7557,41) (9.3920,61) (12.5016,53) (12.1871,42) (15.3495,45)
    (15.0693,59) (16.1583,44) (10.3043,46) (9.4993,33) (8.5735,41) (10.6791,32) (12.1863,45) (8.7999,43)
    (9.1007,33) (9.9696,27) (9.4026,50) (10.3313,42) (10.0131,41) (10.2674,34) (11.7720,60) (10.7328,40)
    (8.9529,35) (8.3663,47) (12.4718,55) (14.7562,36) (14.4323,19) (24.4509,42) (23.4458,25) (19.6099,48)
    (11.6648,60) (17.4452,52) (7.4275,30) (16.6965,55) (8.9976,56) (20.0167,44) (11.1989,46) (15.8012,38)
    (18.0965,40) (14.7693,41) (13.7324,66) (18.7170,60) (12.8484,82) (12.0525,45) (15.7857,62) (10.1660,66)
    (11.4969,64) (7.9753,36) (9.9801,44) (12.7696,45) (10.5710,49) (11.2904,58) (9.7898,58) (9.6853,52)
    (11.2426,54) (15.0957,44) (8.1609,63) (10.6880,76) (9.1074,43) (20.6437,119) (12.3261,62) (11.9265,54)
    (10.4075,56) (10.1166,57) (8.3510,68) (13.2226,62) (6.4457,45) (9.2122,67) (2.4720,23) (3.5962,20)
    (3.0445,31) (9.6385,76) (9.6684,56) (4.3115,29) (11.3884,50) (8.3270,62) (9.5872,80) (13.0065,92)
    (13.2532,72) (10.6245,68) (4.9160,33) (4.1559,19) (8.4825,49) (11.9678,48) (3.8713,34) (8.8973,53)
    (11.4599,58) (10.4725,69) (8.8012,59) (2.9782,20) (7.9717,34) (8.5860,21) (12.2430,71) (9.2324,71)
    (10.3595,54) (12.1710,51) (9.4144,63) (8.4135,77) (12.0937,54) (1.2665,12) (10.8112,68) (8.9098,49)
    (14.6371,79) (8.0279,72) (9.6924,39) (3.1789,19) (4.2691,15) (7.9637,66) (13.7643,77) (9.4965,48)
    (11.4426,75) (8.6003,57) (8.0885,59) (3.0768,37) (12.0952,86) (9.1742,59) (10.5867,63) (9.6430,76)
    (9.6037,52) (9.5926,37) (0.9094,20) (14.7263,57) (8.4936,48) (8.6349,45) (12.0813,53) (8.4952,65)
    (8.5020,87) (9.3738,53) (1.8823,22) (8.7929,83) (1.6342,20) (8.3464,59) (7.9508,54) (9.0516,74)
    (0.8969,24) (8.4467,63) (10.0752,58) (8.4907,84) (11.0561,81) (10.3357,45) (8.4247,74) (12.1664,85)
    (16.3835,84) (11.0919,65) (3.2177,21) (1.1858,12) (3.1665,28) (10.0730,60) (3.5806,32) (6.4814,76)
    (9.9973,53) (7.9692,58) (8.2468,60) (11.5988,68) (8.6129,48) (11.4948,55) (12.4725,50) (11.9362,55)
    (6.8158,44) (8.8999,33) (8.4884,44) (9.6695,61) (10.7106,81) (15.0289,66) (13.7205,55) (8.5532,56)
    (10.6127,43) (9.3694,85) (11.3792,50) (11.0996,59) (8.6135,49) (9.5076,71) (14.5936,56) (13.2314,39)
    (9.1950,53) (13.9814,63) (8.9291,46) (3.7571,30) (9.2450,46) (13.1994,50) (9.6875,43) (7.8067,46)
    (7.2621,65) (9.5968,90) (9.0529,64) (10.8203,67) (15.5284,65) (11.0568,67) (10.2347,83) (8.8978,83)
    (8.2523,56) (8.8677,64) (10.2382,89) (10.8798,92) (12.1661,65) (11.9170,74) (8.3445,33) (10.8196,60)
    (13.1607,59) (8.5890,54) (10.9408,47) (4.2275,23) (7.8140,45) (8.8806,56)
};
\addplot[black, densely dashed, line width=0.9pt, forget plot] coordinates {(0,41.1353) (30,66.8403)};
\node[draw=black!30, fill=white, fill opacity=0.95, text opacity=1,
      rounded corners=1pt, inner sep=1.8pt, font=\fontsize{7.5}{8.5}\selectfont\bfseries]
  at (rel axis cs:0.18,0.88) {$r = 0.29$};
\end{axis}

\begin{axis}[
    at={(path.outer south west)},
    anchor=north west,
    yshift=-0.12cm,
    width=17.5cm, height=3.2cm,
    hide axis, xmin=0, xmax=1, ymin=0, ymax=1,
    legend style={
        at={(0.5,0.5)}, anchor=center,
        legend columns=9,
        font=\fontsize{6.5}{7.5}\selectfont,
        draw=none,
        /tikz/every even column/.append style={column sep=0.55em},
    },
    legend cell align=left,
]
\addlegendimage{area legend, fill=blue!55!black!55, draw=none, opacity=0.85}
\addlegendentry{Histogram}
\addlegendimage{line legend, color=blue!25!black, line width=1.1pt}
\addlegendentry{KDE}
\addlegendimage{line legend, orange!80!red, line width=0.9pt}
\addlegendentry{Median}
\addlegendimage{line legend, red!70!black, densely dashed, line width=0.9pt}
\addlegendentry{Mean}
\addlegendimage{only marks, mark=*, mark size=1.6pt, blue!55!black!70}
\addlegendentry{Casual}
\addlegendimage{only marks, mark=*, mark size=1.6pt, orange!80!red!70}
\addlegendentry{Fine}
\addlegendimage{only marks, mark=*, mark size=1.6pt, green!55!black!55}
\addlegendentry{Formal}
\addlegendimage{only marks, mark=*, mark size=1.6pt, purple!55!black!55}
\addlegendentry{Natural}
\addlegendimage{line legend, black, densely dashed, line width=1.0pt}
\addlegendentry{Trend}
\end{axis}
\end{tikzpicture}%
}
\vspace{-3.0mm}
\caption{Path and instruction statistics for \textbf{HumanoidVLN} ($n=933$).
\textbf{(a--b)} Path- and instruction-length distributions.
\textbf{(c)} Path versus instruction length across four styles ($r=0.29$).}
\label{fig:dataset_statistics}
\vspace{-3.5mm}
\end{figure*}

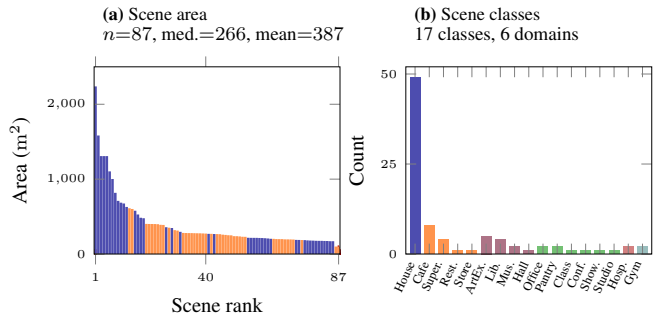
\begin{figure}[!b]
\centering
\resizebox{\columnwidth}{!}{%
\begin{tikzpicture}
\begin{axis}[
    name=area,
    width=3.55cm, height=2.7cm, scale only axis,
    ybar, bar width=0.9, bar shift=0pt,
    enlarge x limits=false,
    xmin=0.5, xmax=87.5, ymin=0, ymax=2500,
    xtick={1,40,87}, ytick={0,1000,2000},
    xlabel={Scene rank}, ylabel={Area ($\mathrm{m}^2$)},
    ylabel near ticks,
    xlabel style={font=\fontsize{8.5}{9.5}\selectfont},
    ylabel style={font=\fontsize{8.5}{9.5}\selectfont},
    tick label style={font=\fontsize{5.0}{5.6}\selectfont},
    title style={font=\fontsize{7.5}{8.5}\selectfont,align=left,
                 at={(0,1)},anchor=south west,yshift=0.5pt},
    title={\textbf{(a)} Scene area\\[-0.5pt]
           {\fontsize{7.8}{8.8}\selectfont $n{=}87$, med.${=}266$, mean${=}387$}},
    axis on top, clip=true,
]
\addplot[red!70!black, densely dashed, line width=0.9pt, forget plot]
  coordinates {(0.5,100) (87.5,100)};
\addplot[ybar, bar width=0.9, bar shift=0pt, fill=blue!55!black!70, draw=none, forget plot] coordinates {
    (1,2238.2857) (2,1582.8465) (3,1307.6802) (4,1307.6798) (5,1307.6798) (6,1103.6683)
    (7,1001.8460) (8,817.2169) (9,709.9143) (10,685.7253) (11,674.8706) (12,628.9261)
    (15,578.1895) (16,528.0540) (17,484.2681) (18,477.9731) (26,357.8017) (28,347.6016)
    (31,296.4821) (41,269.5757) (43,267.0738) (55,217.1348) (56,216.0937) (57,215.9361)
    (58,215.9361) (59,214.5651) (60,213.4233) (61,210.9856) (62,207.7840) (63,203.6400)
    (72,188.9425) (73,188.0859) (75,185.6503) (76,182.3101) (77,180.1055) (78,178.9336)
    (79,176.8243) (80,176.0986) (81,175.9847) (82,174.5644) (83,174.0766) (84,171.3158)
    (85,171.1429)
};
\addplot[ybar, bar width=0.9, bar shift=0pt, fill=orange!80!red!70, draw=none, forget plot] coordinates {
    (13,611.5200) (14,601.7100) (19,402.5100) (20,401.3100) (21,400.9800) (22,400.3400)
    (23,398.4000) (24,390.3200) (25,388.6000) (27,351.5100) (29,318.9900) (30,313.2200)
    (32,281.9300) (33,280.5200) (34,279.6100) (35,279.3900) (36,278.3600) (37,277.2700)
    (38,276.6600) (39,274.8700) (40,273.0000) (42,269.5600) (44,266.3000) (45,265.0300)
    (46,258.4900) (47,255.3500) (48,251.0700) (49,249.6000) (50,238.4100) (51,238.0500)
    (52,235.1900) (53,230.7200) (54,229.7500) (64,202.7500) (65,199.9500) (66,199.0000)
    (67,199.0000) (68,195.8100) (69,195.6900) (70,194.9200) (71,193.8700) (74,187.0800)
    (86,100.0000) (87,100.0000)
};
\end{axis}

\begin{axis}[
    name=cls,
    at={(area.south east)}, anchor=south west, xshift=0.95cm,
    width=3.55cm, height=2.7cm, scale only axis,
    ymin=0, ymax=52, xmin=0.35, xmax=17.65,
    ytick={0,25,50},
    xtick={1,...,17},
    xticklabels={House,Cafe,Super.,Rest.,Store,ArtEx.,Lib.,Mus.,Hall,Office,Pantry,Class,Conf.,Show.,Studio,Hosp.,Gym},
    x tick label style={font=\fontsize{5.2}{5.6}\selectfont, rotate=65, anchor=east},
    y tick label style={font=\fontsize{5.2}{5.8}\selectfont},
    ylabel={Count},
    ylabel near ticks,
    ylabel style={font=\fontsize{8.5}{9.5}\selectfont},
    title style={font=\fontsize{7.5}{8.5}\selectfont,align=left,
                 at={(0,1)},anchor=south west,yshift=0.5pt},
    title={\textbf{(b)} Scene classes\\[-0.5pt]
           {\fontsize{7.8}{8.8}\selectfont 17 classes, 6 domains}},
    axis on top, clip=false,
]
\fill[blue!55!black!70] (axis cs:0.62,0) rectangle (axis cs:1.38,49);
\fill[orange!80!red!70] (axis cs:1.62,0) rectangle (axis cs:2.38,8);
\fill[orange!80!red!70] (axis cs:2.62,0) rectangle (axis cs:3.38,4);
\fill[orange!80!red!70] (axis cs:3.62,0) rectangle (axis cs:4.38,1);
\fill[orange!80!red!70] (axis cs:4.62,0) rectangle (axis cs:5.38,1);
\fill[purple!55!black!55] (axis cs:5.62,0) rectangle (axis cs:6.38,5);
\fill[purple!55!black!55] (axis cs:6.62,0) rectangle (axis cs:7.38,4);
\fill[purple!55!black!55] (axis cs:7.62,0) rectangle (axis cs:8.38,2);
\fill[purple!55!black!55] (axis cs:8.62,0) rectangle (axis cs:9.38,1);
\fill[green!55!black!55] (axis cs:9.62,0) rectangle (axis cs:10.38,2);
\fill[green!55!black!55] (axis cs:10.62,0) rectangle (axis cs:11.38,2);
\fill[green!55!black!55] (axis cs:11.62,0) rectangle (axis cs:12.38,1);
\fill[green!55!black!55] (axis cs:12.62,0) rectangle (axis cs:13.38,1);
\fill[green!55!black!55] (axis cs:13.62,0) rectangle (axis cs:14.38,1);
\fill[green!55!black!55] (axis cs:14.62,0) rectangle (axis cs:15.38,1);
\fill[red!65!black!50] (axis cs:15.62,0) rectangle (axis cs:16.38,2);
\fill[teal!70!black!40] (axis cs:16.62,0) rectangle (axis cs:17.38,2);
\end{axis}
\end{tikzpicture}%
}
\vspace{-3.0mm}
\caption{Scene statistics for \textbf{HumanoidVLN}.
\textbf{(a)} Navigable area of 87 scenes (blue: artist-designed; orange: GS; dashed: $100\,\mathrm{m}^2$).
\textbf{(b)} Counts of 17 indoor classes in six domains.}
\label{fig:scene_diversity}
\vspace{-3.5mm}
\end{figure}

Fig.~\ref{fig:dataset_statistics} and Fig.~\ref{fig:scene_diversity} summarize the benchmark. Across 933 episodes, paths have a median length of 9.97\,m and instructions contain a median of 50 words. The modest path--instruction correlation ($r=0.29$) indicates that instruction length is not determined solely by trajectory length. The 87 scenes span 17 indoor classes across six application domains, with a median navigable area of 266\,m$^2$.

\section{CONCLUSION}
We presented \textbf{HumanoidVLN}, a physics-grounded simulator and benchmark
for VLN across diverse humanoid embodiments, addressing three critical gaps:
embodiment-diverse simulation, navigability-curated scenes, and reliable
grounded instruction generation. Our Isaac Sim platform supports Unitree G1,
Unitree H1, and two internal robots via a hierarchical RL-plus-path-tracker
control stack. The 87-scene suite spans 17 indoor classes curated for
bipedal traversability, and 933 episodes—each with one spatially grounded fine-grained instruction and three coarse-grained stylistic variants—are generated by a \textit{Dual Generator-Reviewer + Paraphraser} MAA with human-in-the-loop verification. Current
limitations include scene diversity, human verification as a scaling
bottleneck, and the computational cost of full physics simulation. Future
work will integrate 3D foundation models for stronger instruction grounding,
develop more efficient 3DGS reconstruction, and expand sim-real validation
on physical humanoid platforms.







\bibliographystyle{ieeetr}   
\bibliography{references}

\end{document}